\pdfoutput=1 

\documentclass[10pt,twocolumn,letterpaper]{article}

\usepackage[pagenumbers]{cvpr} 

\usepackage{graphicx}
\usepackage{amsmath}
\usepackage{amssymb}
\usepackage{booktabs}

\usepackage[natbib=true,style=ieee, maxcitenames=1, mincitenames=1]{biblatex} 
\usepackage[breaklinks,colorlinks]{hyperref}

\usepackage[capitalize]{cleveref}
\crefname{section}{Sec.}{Secs.}
\Crefname{section}{Section}{Sections}
\Crefname{table}{Table}{Tables}
\crefname{table}{Tab.}{Tabs.}

\def\cvprPaperID{} 
\def\confName{}
\def\confYear{}

\author{Carsten T. L\"uth$^{1,2}$\thanks{Work performed at the Division of Medical Image Computing$^3$} , 
David Zimmerer$^{3}$, 
Gregor Koehler$^{3}$,
Paul F. Jaeger$^{1,2}$\\
Fabian Isensee$^{2,3}$,
Jens Petersen$^{3}$,
Klaus H. Maier-Hein$^{3}$
\smallskip
\\ 
$^1$Interactive Machine Learning Group,
German Cancer Research Center (DKFZ)
\\
$^2$Helmholtz Imaging, 
German Cancer Research Center (DKFZ)
\\
$^3$Division of Medical Image Computing, 
German Cancer Research Center (DKFZ)
\smallskip
\\
{\tt\small carsten.lueth@dkfz-heidelberg.de}
}

\usepackage{multirow}

\title{CRADL: Contrastive Representations for Unsupervised Anomaly Detection and Localization}

\begin{document}

\maketitle
\begin{abstract}
    Unsupervised anomaly detection in medical imaging aims to detect and localize arbitrary anomalies without requiring annotated anomalous data during training. 
Often, this is achieved by learning a data distribution of normal samples and detecting anomalies as regions in the image which deviate from this distribution.
Most current state-of-the-art methods use latent variable generative models operating directly on the images.
However, generative models have been shown to mostly capture low-level features, s.a. pixel-intensities, instead of rich semantic features, which also applies to their representations.
We circumvent this problem by proposing CRADL whose core idea is to model the distribution of normal samples directly in the low-dimensional representation space of an encoder trained with a contrastive pretext-task.
By utilizing the representations of contrastive learning, we aim to fix the over-fixation on low-level features and learn more semantic-rich representations.
Our experiments on anomaly detection and localization tasks using three distinct evaluation datasets show that 1) contrastive representations are superior to representations of generative latent variable models and 2) the CRADL framework shows competitive or superior performance to state-of-the-art.
\end{abstract}

\section{Introduction}
\label{s:intro}
The task of anomaly detection is a long-standing task in the medical domain, with many diseases being defined by their deviation from what is considered to be normal \cite{ZYTKOWSKI2021100105}.
In medical image analysis detecting and localizing anomalies is, therefore also often the general goal.
With powerful deep learning algorithms, this task is often tackled using supervised machine learning, which can be highly effective given that enough diseased cases are available and annotated during the training process \cite{aggarwal2021diagnostic, mckinneyInternationalEvaluationAI2020}.
However, most supervised models are not explicitly designed to handle Out-of-Distribution (OoD) data. They thus might struggle to extrapolate to new settings e.g., the heterogeneity of diseases beyond the training distribution \cite{kellyKeyChallengesDelivering2019}. 
Consequently, each new class of pathology or imaging modality necessitates the creation of new annotated datasets---a process that scales poorly with the large number of existing pathologies and the ever-increasing amount of image acquisition methods.
Unsupervised anomaly detection promises to deliver predictions in the absence of annotated diseased data.
Thus, overcoming the need for cumbersome manual annotations of diseased images, this class of methods could offer a far greater breadth of applications. 
In principle, this can be realized by learning the distribution of healthy samples. Images (or rather some voxels in the images) `deviating' from this distribution are then defined as outliers. The problem of detecting these deviations can be posed as an OoD detection problem similar to the statistical tests underlying  blood tests using reference ranges \cite{bainDacieLewisPractical2017}.
Specifically, in the medical imaging domain, the current state-of-the-art methods for anomaly detection are latent variable generative models operating directly on image space, mainly different subtypes of and scoring methods based on Variational Autoencoders (VAEs) and Generative Adversarial Networks (GANs) \cite{baurAutoencodersUnsupervisedAnomaly2020,chenUnsupervisedLesionDetection2020a,schleglFAnoGANFastUnsupervised2019}.
It has been shown, however, that the likelihoods of generative models tend to focus on low-level features such as background characteristics 
\cite{nalisnickDeepGenerativeModels2019,renLikelihoodRatiosOutofDistribution2019,xiaoLikelihoodRegretOutofDistribution2020,meissen2021challenging} and that their representations have problems capturing semantic information \cite{nalisnickDeepGenerativeModels2019,zimmererContextencodingVariationalAutoencoder2018a}. 
Therefore the anomaly scores of these methods built on their likelihoods and representations  are heavily dependent on the aforementioned characteristics not capturing fine semantic information. 
\newpage
\textit{
Driven by the hypothesis that more discriminative semantic features are superior to low-level features, we investigate if the representations obtained with self-supervised contrastive learning can aid unsupervised anomaly detection and localization.}

For this, we propose CRADL, a simple unsupervised representation-based OoD framework consisting of a generative model working on the representations of a feature extractor, trained in a self-supervised fashion based on SimCLR \cite{chenSimpleFrameworkContrastive2020}.

Our experimental results comparing it with state-of-the-art latent generative models and using the representation-based approach with generative representations on MRI-Brain lead to the following findings:
\begin{itemize}
    \item In the low-dimensional representation space of CRADL (and generative models) computationally inexpensive generative models allow for reliable detection and even localization of anomalies.
    \item CRADL is comparable with state-of-the-art anomaly detection and localization approaches and is better suited to detect fine deviations from normality.
    \item Representations obtained with the contrastive pretext task of CRADL outperform representations of our latent generative models.
\end{itemize}
The remainder of this work is structured as follows: In \cref{s:improved-task}, we give a brief task description and explain the thought process based on the literature landscape motivating our purely representation-based approach for anomaly detection. 
In \cref{s:methods} we detail the way CRADL obtains and uses representations allowing for anomaly detection and localization.
Following this description, we give our experimental details in \cref{s:expsetup} and set the results into the context of our findings in \cref{s:results}.
Finally, we conclude the advantages of CRADL in \cref{s:conclusion} with potential use cases based on its properties and the results of our experiments.

\section{Improved Anomaly Detection \& Localization}
\label{s:improved-task}
\subsection{Anomaly Detection \& Localization from an Out-of-Distribution Perspective}
In \textbf{anomaly and OoD detection} we are interested in differentiating between normal samples (inliers) $x$ of a Dataset $\mathcal{D}$, drawn from the underlying distribution $p^*(x)$, and any anomalous samples, termed outliers $x_{\mathrm{ano}}$, which are drawn from a different --but generally unknown-- distribution $Q(x)$.
In OoD detection differentiating between inliers and outliers is commonly tackled by simply defining a score-based binary classifier with the help of a threshold. 
If $Q(x)$ were known exactly, the most straightforward approach would be to define the Bayes Optimal Score
\begin{equation}
   s_B(x) = -\log(\dfrac{p^*(x)}{Q(x)}).
\end{equation}
Generally, however, there is a distinct lack of anomalous samples drawn from $Q(x)$, which is made even more complicated due to the fact that the distribution $Q(x)$ is implicitly defined  by the selection of outliers, which can be done arbitrarily. E.g., $Q(x)$ can be data stemming from a different dataset, noisy images from the original dataset,  or samples containing (previously unknown) classes. \\ 
To circumvent this problem, a common approach is to use the NLL score $s^*(x) = - \log(p^*(x))$ where a generative model is used to approximate the underlying distribution $p(x) \approx p^*(x)$. 
This score can then detect outliers based on the assumption that generally: $s(x) \leq s(\hat{x})$ for $x \sim p^*(x)$, $x_{\mathrm{ano}} \sim Q(x)$.
\footnote{In words: The probability of inliers is higher than for outliers under the inlier distribution.}

For \textbf{anomaly localization} one wants to know now which regions of a singular sample $x$ are deviating from normality. 
Based on this task description, the implicit assumption for anomaly localization is that an anomalous sample $x_{\mathrm{ano}}$ is somewhat similar to a normal sample $x_{\mathrm{norm}}$ in a manner where $x_{\mathrm{ano}} = x_{\mathrm{norm}} + a$. Here, $a$ is the change making the sample anomalous.
From an OoD perspective, this might be interpreted as raising the question: \textit{If this sample was normal, where would it differ?}
The detection of $a$ is tackled in the form of an anomaly heatmap $l(x)$ with the same spatial dimension as $x$ which can either directly be learned or be based on the OoD detection function $l(x) = \text{Localize}(s(x))$.
The goal of $l(x)$ is to indicate regions of an anomalous sample where $a$ leads to changes from the normal samples while not highlighting any areas for normal samples.
\subsection{Moving from Image Space to Representations}
Current state-of-the-art for anomaly detection and localization on medical images are generative methods, s.a., VAEs and GANs operating directly in image space. 
The most forward and commonly used approach is the reconstruction difference scoring for both detection and localization \cite{zimmererUnsupervisedAnomalyLocalization2019,baurAutoencodersUnsupervisedAnomaly2020}.
For anomaly detection, it has been shown to be advantageous to also use the information in the representation space for anomaly detection \cite{zimmererUnsupervisedAnomalyLocalization2019,zimmererContextencodingVariationalAutoencoder2018a,baur2018deep}.
\citet{baurAutoencodersUnsupervisedAnomaly2020} compared the most common methods based on their anomaly localization capabilities.
They conclude that a VAE-based iterative image restoration setting \cite{chenUnsupervisedLesionDetection2020a} making heavy use of the learned representations performed best across most datasets they evaluated. 
A similar trend has been observed by \citet{zimmererContextencodingVariationalAutoencoder2018a} where they show that the backpropagated loss of the KL-Divergence purely using the representations improved anomaly localization capability.
Other lines of work using VQ-VAEs with discrete latent spaces capturing spatial information explicitly in conjunction with generative models on the latent space showed 
that a restoration approach purely using representations for resampling of unlikely regions improved anomaly localization performance \cite{pinayaUnsupervisedBrainAnomaly2021,marimont2021anomaly}.
\textbf{Take Away:} Actively using the distribution in the representation space of generative models seems to improve the anomaly detection and localization performance of generative methods applied on image space.
\paragraph{General Problems of Generative Models on Image Space}
The image space is highly dimensional, leading to two naturally arising problems of generative models, which are the curse of dimensionality --inherent to probability distributions--
\footnote{The curse of dimensionality states that the highest likelihood region in a high dimensional space is basically devoid of samples.}
and the number of samples needed for a reliable fit of generative models --drastically increasing with the complexity of the model and the dimensionality of the distribution.
However, the distribution of images in the image space is very complex (e.g., non-Gaussian) and is hypothesized to live on a very thin manifold of the whole space.

These properties might be one of the main reasons why it has been shown several times in natural imaging that generative models on pixel space have a hard time detecting samples being deemed as OoD \cite{nalisnickDeepGenerativeModels2019,nalisnickDetectingOutofDistributionInputs2019}.
For example, a generative model trained on CIFAR10 gives samples from the SVHN generally a higher likelihood than samples on CIFAR10 \cite{nalisnickDeepGenerativeModels2019,nalisnickDetectingOutofDistributionInputs2019}.
It is hypothesized that this might be due to the curse of dimensionality where SVHN is --from a pixel perspective--, basically, a subset of CIFAR-10 in the empty high-likelihood region \cite{nalisnickDetectingOutofDistributionInputs2019}.
Other works fit a generative model on the low-dimensional representations of supervised classifiers, showing that even simple generative models allow solving the OoD problem mentioned above reliably if used on semantically rich representations\cite{liangEnhancingReliabilityOutofdistribution2020,leeSimpleUnifiedFramework2018,winkensContrastiveTrainingImproved2020,zhangHybridModelsOpen2020}.
\textbf{Take Away:} Lower-dimensional representations carrying meaningful information alleviate inherent problems of generative models on high-dimensional image spaces regarding fit and OoD detection ability.
\paragraph{The Problem of Obtaining and Usefulness of Meaningful Representations}
For standard medical tests, e.g., a blood count, meaningful representations to model the distribution of the healthy population are features s.a. hemoglobin and white blood cell levels, which are directly read out by a procedure \cite{bainDacieLewisPractical2017}.
In general, obtaining meaningful features for medical imaging and images is not as straightforward.
Deep Learning is often seen as a representation learning task, and latent variable generative models were historically used for the purpose of representation  learning \cite{Goodfellow-et-al-2016}.
Currently, supervised learning is the de-facto standard for  down-stream tasks on images, s.a. classification also leading to the best results, but it requires considerable amounts of annotated training data of each class \cite{chenSimpleFrameworkContrastive2020}. 
As noted in the paragraph above, supervised representations have been shown to work well in the context of OoD with works using generative models on the representations of models trained in a supervised fashion.
However, in cases where there is not sufficient annotated training data and/or the task associated with the annotation does not require capturing information, potentially crucial information about a sample is lost since injectivity is not per se a requirement which is a possible explanation for why classifiers are often overconfident for OoD inputs \cite{winkensContrastiveTrainingImproved2020}.
Alternatively, the representations of a generative model could be used, which, however, have been shown to focus on encoding low-level features, s.a., background and color on natural images leading to blurry reconstructions and generations \cite{maaloeBIVAVeryDeep2019}.
Recently Self-Supervised Learning and, more specifically, contrastive learning have shown to be able to learn rich semantic representations without requiring label information by drastically reducing the amount of labeled data to reach fully-supervised down-stream task classification performance, substantially outperforming generative approaches in this setting. \cite{chenBigSelfSupervisedModels2020,grillBootstrapYourOwn2020}.
\textbf{Take Away:} Self-Supervised Learning could lead to representations better suited for anomaly detection than that of generative models and also reduce the number of explicitly known annotated healthy samples compared to a fit of a generative model on image space.

\section{Methods}
\label{s:methods}
\begin{figure*}[tb]
    \centering
    \includegraphics[width=.70\textwidth]{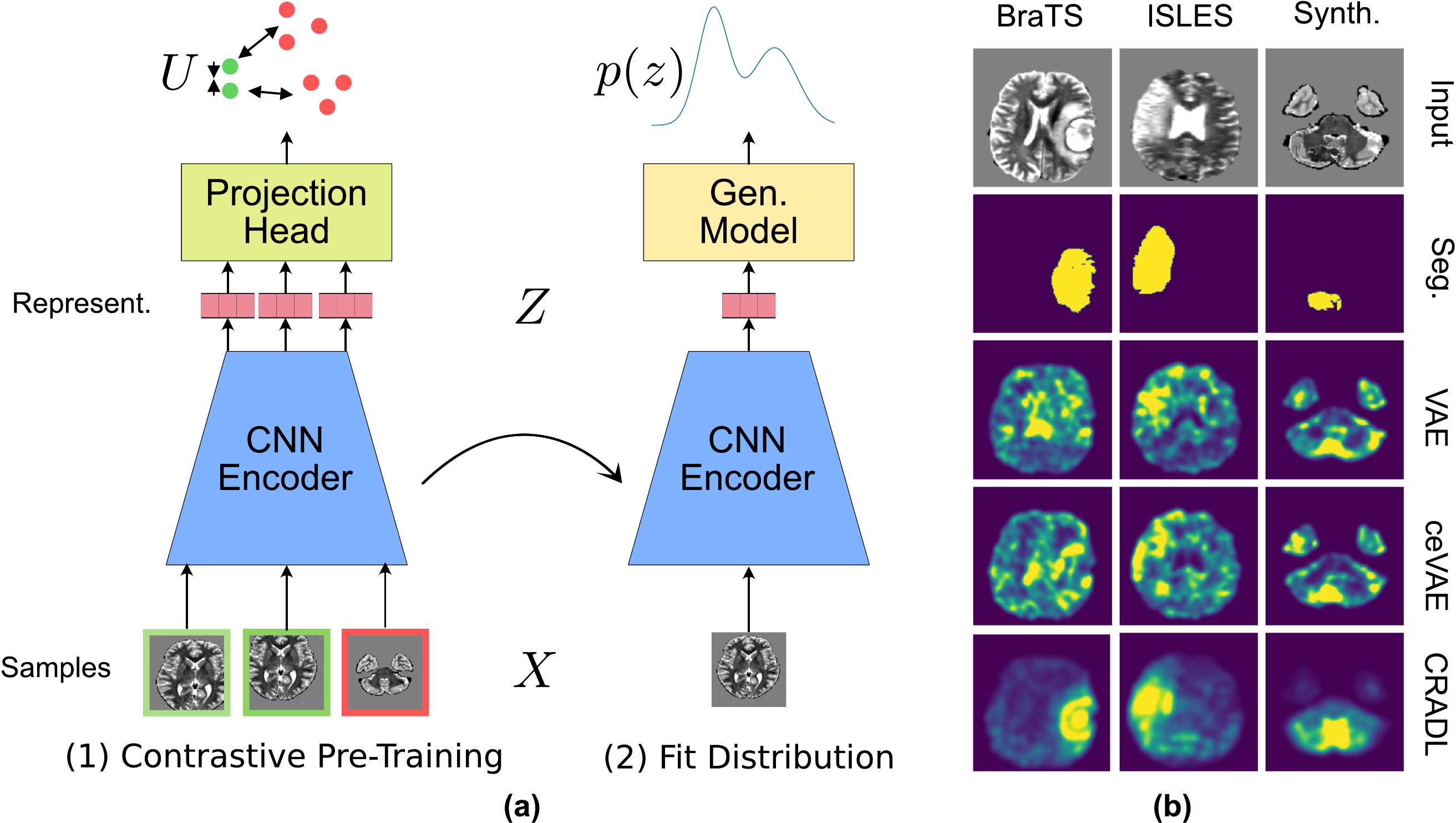}
    \caption{(a) Visualization of the fitting pipeline, from contrastive pretext (SimCLR) (1) to fitting of the generative model (2). (b) Pixel-wise scores - clamped and normalized for visual inspection.}
    \label{fig:fit_pipeline}
\end{figure*}

We propose CRADL, a method using \textit{Contrastive Representations for unsupervised Anomaly Detection and Localization}. 
Our proposed approach is comprised of two stages, as shown in \cref{fig:fit_pipeline}. 
During the first stage, the encoder $f$, which maps a sample $x$ from the image space $X$ to a learned representation space $Z$ : $z=f(x)$, is trained in a self-supervised fashion with a contrastive task.
The contrastive task should lead to the representations that are clustered based on semantic similarity which is defined implicitly by the task.
In the second stage these clustered representations of the normal samples are used to fit a generative model $p(z)$. 
The anomaly-score of a sample $x$ is then given by the negative-log-likelihood (NLL) of its representation
\begin{equation}
    s(x) = - \log (p(f(x))).
\end{equation}
Anomaly localization is then performed using pixel-level anomaly scores (anomaly heatmap) for a sample by back-propagating the gradients of the NLL of the representation of a sample into the image space
\begin{equation}
    l(x) = \left|\Delta_x \log s(x)\right|.
\end{equation}
Implicitly this approach assumes that regions with large gradients exhibit anomalies.

\subsection{Contrastive Training}
\label{ss:methods-contrastive}
During the first stage of training CRADL, a contrastive pretext-task is used to train the encoder.
The contrastive task we use is conceptually identical to SimCLR \cite{chenSimpleFrameworkContrastive2020} but uses different data augmentations changed for the medical domain as transformations.
When using SimCLR, positive pairs are obtained using data augmentations $t$ drawn randomly from a set of augmentations $\mathcal{T}$ and all other pairs of the samples are used as negatives. 
To this end, each sample $x_i$ in a minibatch of $N$ examples is transformed twice, yielding two different views that make up the positive pair. 
The representations produced by feeding the views through the encoder $f$ and projection head $g$, $\tilde{u}_{i} = g(f(\tilde{t}(x_i)))$ and $\hat{u}_{i} = g(f(\hat{t}(x_i)))$, are encouraged to be similar by optimizing the NT-Xent contrastive loss: 
\begin{equation}
    L(x_i) = - \log \dfrac
    {\exp(\text{sim}(\tilde{u}_{i}, \hat{u}_{i}) / \tau)}
    {
    \sum_{\bar{u} \in \Lambda^{-}} \exp(\text{sim}(\tilde{u}_{i}, \bar{u})) / \tau) }
\end{equation}
Here the set $\Lambda^{-}$ are the negative pairs for a sample $x_i$ consisting of all examples except $\tilde{u}_i$, which are all other $2N-1$ examples in the minibatch. 
The similarity measure $\text{sim}(x,y)$ is cosine similarity.
The loss over the whole minibatch is then obtained by summing over all positive pairs.

\subsection{Generative Model}
\label{ss:methods-generative}
The generative model operating on the representation space to detect and localize anomalies, which is trained in the second stage, can be essentially any generative model. In practice, there just needs to be a selection or aggregation criterion.
However, due to the encoder's clustered representations, a `simple' generative model can be used. 
Concretely we decide to use two families being:

\noindent\textbf{Gaussian Mixture Model} A Gaussian Mixture Model (GMM) is used since it is one of the simplest generative models. 
The probability distribution of a GMM with $K$ components is noted in equation \ref{eq:gmm}. We fit the GMM with the Expectation-Maximization (EM) Algorithm \cite{dempsterMaximumLikelihoodIncomplete1977} with $K$ being the number of components (specified before the fit). 
\begin{equation}\label{eq:gmm}
    p(z; \Theta) = \sum_{k=1}^K \mathcal{N}(z; \mu_k, \Sigma_k) \cdot \pi_k
\end{equation}
Here $\mu_k$ is the mean, $\Sigma_k$ is the covariance matrix and $\pi_k$ is the bias of mode $k$.

\noindent\textbf{Normalizing Flow}
A simple Normalizing Flow ($\text{Flow}$) based on the invertible and volume-preserving RealNVP architecture \cite{dinhDensityEstimationUsing2017a} is used since it allows to model complex probability distributions.
This is achieved through the use of the change of variable formula in combination with a standard normal (Gaussian) $\mathcal{N}(0,\text{I})$ with the same dimensionality as $Z$ as a prior in its latent space 
\begin{equation}
    p(z; \Theta) = \mathcal{N}(\text{Flow}(z); 0, \text{I}) \cdot \left|\det\left( \text{Jacobian}_{\text{Flow}}(z) \right) \right|.
\end{equation}
The Normalizing Flow should be able to give an upper limit to the performance achievable with GMMs with more components.

\section{Experimental Setup}
\label{s:expsetup}
All experiments were performed three times with different seeds to obtain a mean and standard deviation explicitly stated for all experimental results.

\subsection{Datasets}
\label{ss:datasets}
All our experiments are conducted on T2-weighted MRI brain datasets consisting of 3D medical images.
All datasets are pre-processed similarly, with the final step being  image-wise z-score normalization of each image and clipping the intensity from -1.5 to 1.5 following \cite{zimmererContextencodingVariationalAutoencoder2018a}.
For use with our models, the 3D images are sliced along the z-axis and are used as 2D inputs, which are then resized to a resolution of 128x128.
This approach to reducing the complexity of the volumetric data and pre-processing is standard procedure \cite{baurAutoencodersUnsupervisedAnomaly2020, zimmererContextencodingVariationalAutoencoder2018a}
\paragraph{Training Dataset}
For training all of our models, we use a subset of the \textbf{HCP} dataset \cite{vanessenHumanConnectomeProject2012}, which purely consists of `normal' MRI Scans, using 894 scans split into training and validation sets.
The HCP dataset consists of young and healthy adults taken in a scientific study.
\paragraph{Anomalous Datasets}
We use three anomalous datasets with test and validation sets for score selection to evaluate the anomaly detection and localization performance, all showcasing a 100\% anomaly prevalence per image based on the presence of a tumor or diseased area in the segmentation mask. 
These three datasets are:

\noindent\textbf{HCP Synth.} An artificially created anomalous dataset we created based on the leftover 98 images from the HCP dataset (not used for training) by rendering real-world objects into brain regions as anomalies following the approach by \citet{zimmererMedicalOutofDistributionAnalysis2020}.
This allows the test set to have the same original distribution (i.e., population, same scanner, site, registration...) as the training set, with only the anomalies differing. 

\noindent\textbf{ISLES} The ischemic stroke lesion segmentation challenge 2015 (ISLES) dataset \cite{maierISLES2015Public2017} with anomalous regions being the stroke lesions consisting of 28 scans taken from 2 institutions during clinical routine. 
This dataset is explicitly stated to contain different pathologies which are not annotated.

\noindent\textbf{BraTS} The Brain Tumor Segmentation Challenge 2017 (BraTS) dataset \cite{bakasAdvancingCancerGenome2017} with anomalous regions being glioblastoma and lower grade glioma consisting of 286 scans taken from multiple institutions during clinical routine.

To allow better interpretation of our results, we give a short dataset description with values for interpreting the results for anomaly detection based on slices and anomaly localization based on voxels can be seen in \cref{tab:data_statistics}.

\begin{table}[h!]
    \centering
    \resizebox{\linewidth}{!}{
\begin{tabular}{l|cc|cccc|cc}
\toprule
Level      & \multicolumn{2}{c|}{Scans}         & \multicolumn{4}{c|}{Slices}                                                          & \multicolumn{2}{c}{Voxels}                      \\
Value      & \multicolumn{2}{c|}{\# in Dataset} & \multicolumn{2}{c}{\# in Dataset} & \multicolumn{2}{c|}{Anomaly Prevalence} & \multicolumn{2}{c}{Anomaly Prevalence} \\
Dataset    & Test            & Val.            & Test             & Val            & Test                   & Val                    & Test                   & Val                   \\
\midrule
HCP Synth. & 49              & 49              & 7105             & 7105           & 20.5\%                   & 19.8\%                   & 0.649\%                   & 0.770\%                 \\
ISLES      & 20              & 8               & 2671             & 1069           & 36.6\%                   & 34.1\%                   & 1.140\%                   & 1.099\%                 \\
BraTS      & 266             & 20              & 35910            & 2700           & 49.0\%                   & 46.0\%                   & 2.427\%                   & 1.923\% \\
\bottomrule
\end{tabular}
}
        \caption{Statistically relevant characteristics of the anomalous datasets used in our evaluation. The baseline AUPRC of a random score is equal to the prevalence.}
    \label{tab:data_statistics}
\end{table}

\paragraph{Distribution Shifts}
Several distribution shifts from the HCP dataset created for scientific purposes to clinical datasets ISLES and BraTS lead to different intensity histograms and prevalences on slice and voxel levels.
Firstly, a population shift is present due to the high prevalence of disease in both clinical datasets which is much higher than that of the whole human population and also often correlated with the age of the population. 
Therefore are not only unannotated anomalies present in images from BraTS and ISLES but also structural changes due to the different ages of the underlying populations of the datasets.
Secondly, a shift in the images is present due to different scanners used to take the images. Generally, the scans for the scientific HCP dataset are taken with an MRI-Scanner with a higher magnetic field strength than used in clinical practice for both BraTS and ISLES.

\subsection{Model}
We use a unified model architecture for our experiments which is based on the deep convolutional architecture from \citet{radfordUnsupervisedRepresentationLearning2016}, so our encoder solely consists of 2D-Conv-Layers and our decoder (for the VAE models) of 2D-Transposed-Conv-Layers (for more details, please refer to the Appendix). 
We chose an initial feature map size of 64 and a latent dimension of 512.
For the projection head of CRADL, we use a simple 2-layer MLP with ReLU non-linearities, a 512 dim. hidden layer, and 256 dim. output.
We performed for both our baselines and our proposed method experiments with different bottleneck sizes and feature map counts.\\

\subsection{Training CRADL}
\paragraph{Contrastive Pretext-Task}
The contrastive pretext training of the encoder is performed for 100 epochs on the HCP training set using the Adam Optimizer, a learning rate of 1e-4, Cosine Annealing \cite{loshchilovSGDRStochasticGradient2017}, 10 Warm-up Epochs and a weight decay of 1e-6, the temperature of the contrastive loss is 0.5 \cite{chenSimpleFrameworkContrastive2020}.
The encoder for later evaluation is selected based on the smallest loss on the HCP validation set.
As transformations for generating different views for the contrastive task, we use a combination of random cropping, random scaling, random mirroring, rotations, and multiplicative brightness, and Gaussian noise.
\paragraph{Generative Model}
We fit a batch of GMMs with $K\in\{1,2,4,8\}$ on representations of the encoder from all samples in the HCP training set without any augmentation. The means of the components are randomly initialized, and the convergence limit for the EM algorithm is set to $0.1$.
In the Appendix, we present the performance of CRADL in relation to $K$. 

The Normalizing Flow is also trained on the representation space in the same manner as the GMM to maximize the likelihood of the samples \cite{dinhDensityEstimationUsing2017a}.
All further information regarding the exact training process and model architecture are shown in the Appendix.

\subsection{Baselines}
We train both the VAE and ceVAE for 100 epochs using the Adam Optimizer, a learning rate of 1e-4, and the unified architecture for both the encoder and decoder on the HCP training dataset. 
The final models for evaluation are chosen based on the lowest loss on the HCP validation set.
The transformations we use during training of the VAE consist of random scaling, random mirroring, rotations,  multiplicative brightness, and Gaussian noise, which have shown clear performance improvements in our early experiments. 
For the ceVAE, we add random cutout transformations \cite{zimmererContextencodingVariationalAutoencoder2018a}.  

\subsection{Inference}
We select the best scoring function for CRADL and all baselines concerning each anomalous evaluation dataset based on the best validation set AUPRC metric (see Appendix).
For CRADL, the scoring functions are the different generative models on its representations. For the VAE and ceVAE, these are the different terms of the loss (Reconstruction difference (Rec.)\cite{baur2018deep}, KL-Divergence \cite{zimmererContextencodingVariationalAutoencoder2018a}, ELBO/combi\cite{zimmererContextencodingVariationalAutoencoder2018a,baur2018deep}).
We deem this approach of scoring function selection reasonable and motivate the choice of generative model selection for CRADL by the fact that the fit of all GMMs and the Normalizing Flow on representations take only 50 minutes on one GPU together compared to more than 8 hours for the fit of one VAE or ceVAE.

\paragraph{\textbf{Post-processing of Anomaly Heatmaps}}
The post-processing pipeline for the pixel-level scores is identical for all methods evaluated and, based on the approach from \citet{baurAutoencodersUnsupervisedAnomaly2020}, restricted to only use one sample (2D): We zero out all pixel scores outside the brain region. In the next step, 2D median pooling (kernel size=5) is applied to filter out edges and single outliers. As the last step, Gaussian smoothing is applied, inspired by the finding of sparse gradients and convolutional artifacts by \citet{zimmererContextencodingVariationalAutoencoder2018a}. Empirically, the reconstruction-based scores of the VAE and ceVAE benefited from this step.

\subsection{Evaluation Metrics}
For the two different tasks of anomaly detection and localization, the discriminative power of the scoring function is measured using the Area Under the Receiver Operator Curve (AUROC) and the Area Under the Precision-Recall Curve (AUPRC) due to their independence of a threshold with anomalies being defined as positives.
In our setting, it is more critical to detect outliers than inliers in combination with a larger number of normal slices/voxels than anomalous slices/voxels. Hence, we emphasize the importance of the AUPRC score because it better captures the detection of anomalies.

\noindent For \textbf{anomaly detection}, we aggregate all slices of a dataset, and each slice is seen as anomalous if more than five voxels contain an anomaly.

\noindent For \textbf{anomaly localization}, we aggregate the anomaly heatmaps of all slices and the corresponding ground-truth segmentations.

\section{Results \& Discussion}
\label{s:results}
First, we analyze the performance of CRADL for anomaly detection and localization with our re-implementations of the state-of-the-art methods VAE and ceVAE \cite{baurAutoencodersUnsupervisedAnomaly2020,zimmererContextencodingVariationalAutoencoder2018a} in sections \ref{ss:results-detection} and \ref{ss:results-localization}.
Then, we analyze the discriminative power of the representations obtained with contrastive learning to that of the aforementioned generative models in the context of anomaly detection and localization in \ref{ss:results-representations}.

The results of every scoring method of each anomaly detection and localization approach are shown in the Appendix, whereas in the following subsections, only relevant results are portrayed.

\subsection{Comparison to State-of-the-Art for Anomaly Detection}
\label{ss:results-detection}
\begin{table}[]
    \centering
    \resizebox{\linewidth}{!}{
\begin{tabular}{l| rr|rr|rr}
\toprule
Method & \multicolumn{2}{c|}{HCP Synth.} & \multicolumn{2}{c|}{ISLES} & \multicolumn{2}{c}{BraTS} \\
       & \multicolumn{1}{c}{AUPRC}          & \multicolumn{1}{c|}{AUROC}         & \multicolumn{1}{c}{AUPRC}       & \multicolumn{1}{c|}{AUROC}       & \multicolumn{1}{c}{AUPRC}       & \multicolumn{1}{c}{AUROC}       \\
\midrule
CRADL  & \bfseries 77.8(0.7)      & \bfseries 89.8(0.3)     & \bfseries 54.9(0.5)   & 69.3(0.0)   & 81.9(0.2)   & 82.6(0.3)   \\
ceVAE  & 43.2(0.2)      & 76.9(0.1)     & 54.1(0.8)   & \bfseries 72.7(0.1)   & \bfseries 85.6(0.2)   & \bfseries 86.5(0.1)   \\
VAE    & 48.5(0.9)      & 79.4(0.1)     & 51.9(0.6)   & 71.7(0.2)   & 80.7(0.7)   & 83.3(0.6)  \\
\bottomrule
\end{tabular}
}
    \caption{Slice-wise anomaly detection performance. AUROC and AUPRC are given in: mean(standard deviation)\% }
    \label{tab:detection-results}
\end{table}
By comparing CRADL with our re-implementations of the VAE and ceVAE for slice-wise anomaly detection, 
we want to assess the general performance of our approach to that of state-of-the-art baselines.
The results are shown in \cref{tab:detection-results}.
For the HCP Synth. dataset CRADL substantially outperforms both the VAE and ceVAE. 
On both ISLES and BRATS, the ceVAE is the best-performing approach showcasing the advantage of the additional self-supervised task on ISLES, whereas it is tied with the VAE on BraTS.

Due to the aforementioned distribution shifts from the normal (healthy) population on HCP to ISLES and BraTS and the more diverse and subtle set of anomalies on HCP Synth. we conclude, therefore, that CRADL can capture fine deviations from normality better than a VAE and ceVAE.
And it can be hypothesized that this exact property of CRADL being able to capture delicate deviations from normality better than a VAE and ceVAE might be the reason for it performing worst on ISLES and BraTS since, based on a distribution perspective, essentially all slices from ISLES and BraTs could be considered anomalous.

This hypothesis could also be backed up since on HCP Synth. the best performing scoring methods are based on the GMM with 8 components and the Normalizing Flow, which both should be able to capture even relatively complex distributions well, while for ISLES and BraTS the GMM with 1 component leads to the best anomaly detection performance where the latter is not as well suited to capture the fine difference in the representation space than the former.
\subsection{Comparison to State-of-the-Art Anomaly Localization}
\label{ss:results-localization}
\begin{table}[]
    \centering
    \resizebox{\linewidth}{!}{
\begin{tabular}{l| rr|rr|rr}
\toprule
Method & \multicolumn{2}{c|}{HCP Synth.} & \multicolumn{2}{c|}{ISLES} & \multicolumn{2}{c}{BraTS} \\
       & \multicolumn{1}{c}{AUPRC}          & \multicolumn{1}{c|}{AUROC}         & \multicolumn{1}{c}{AUPRC}       & \multicolumn{1}{c|}{AUROC}       & \multicolumn{1}{c}{AUPRC}       & \multicolumn{1}{c}{AUROC}       \\
\midrule
CRADL  & \bfseries 32.5(0.8)      & \bfseries 97.8(0.0)     & \bfseries 18.6 (3.9)  & \bfseries89.8(0.3) & 38.0(1.6)  & 94.2(0.1)  \\
ceVAE  & 17.2(1.5)      & 92.1(0.4)     & 14.5(1.3)  & 87.9 (0.2)  & \bfseries 48.3(3.0)  & \bfseries 94.8(0.3)  \\
VAE    & 24.9(0.6)      & 95.5(0.1)     & 7.7(1.2)   & 87.5 (0.6)  & 29.8(0.4)  & 92.5(0.1) \\
\bottomrule
\end{tabular}
}
    \caption{Voxel-wise anomaly localization performance. AUROC and AUPRC are given in: mean(standard deviation)\% }
    \label{tab:all_results}
\end{table}
The goal of this experiment is to verify the effectiveness of our approach for anomaly localization by backpropagating gradients of the NLL of generative models operating on a representation space to the image space through an encoder.
To this end, we compare our results again with our state-of-the-art implementations of a VAE and ceVAE each with three different scores, which are reconstruction differences, backpropagated gradients of the KL-Divergence \cite{zimmererContextencodingVariationalAutoencoder2018a} and the combi-scoring based on a multiplication of KL-Divergence gradients and reconstruction differences \cite{zimmererContextencodingVariationalAutoencoder2018a}.
The results of the best-performing anomaly localization scoring methods for CRADl, VAE and ceVAE are shown in \cref{tab:all_results}.
On HCP Synth. and ISLES CRADL outperforms both VAE and ceVAE. On the BraTS dataset, the ceVAE is the best-performing method followed by the VAE.
The trend already discussed in \cref{ss:results-detection} continues where HCP Synth. favors for CRADL more complex generative models with the performance increasing, for higher values of K.
Interestingly the Normalizing Flow does not perform well, but this might be due to convoluted gradients in the representation space, which do not behave as `nice' as the one of a GMM approximately scaling quadratically with regard to Mahalanobis distance from a center. 
The trend that using representations for anomaly localization is advantageous can be observed for the VAE and ceVAE where the combi-score performs best by combining both information from the representation and the prior distribution with the reconstruction error by defining a distribution in image space \cite{zimmererContextencodingVariationalAutoencoder2018a}.  
Further, the fact that CRADL performs not as much better on ISLES and worse on BraTS than the ceVAE might be due to the context-encoding task better suited to these datasets where anomalies regions are lesions that often glow brightly in T2-weighted MRI scans.

We interpret the strong results on HCP Synth for anomaly localization to showcase that CRADL can localize anomalies that are close to the normal distribution and diverse in appearance (not only bright glowing) better than both ceVAE and VAE.

\subsection{Anomaly Detection \& Localization with Generative Models on Representation}
\label{ss:results-representations}
\begin{figure}[t]
    \centering
    \includegraphics[width=\linewidth]{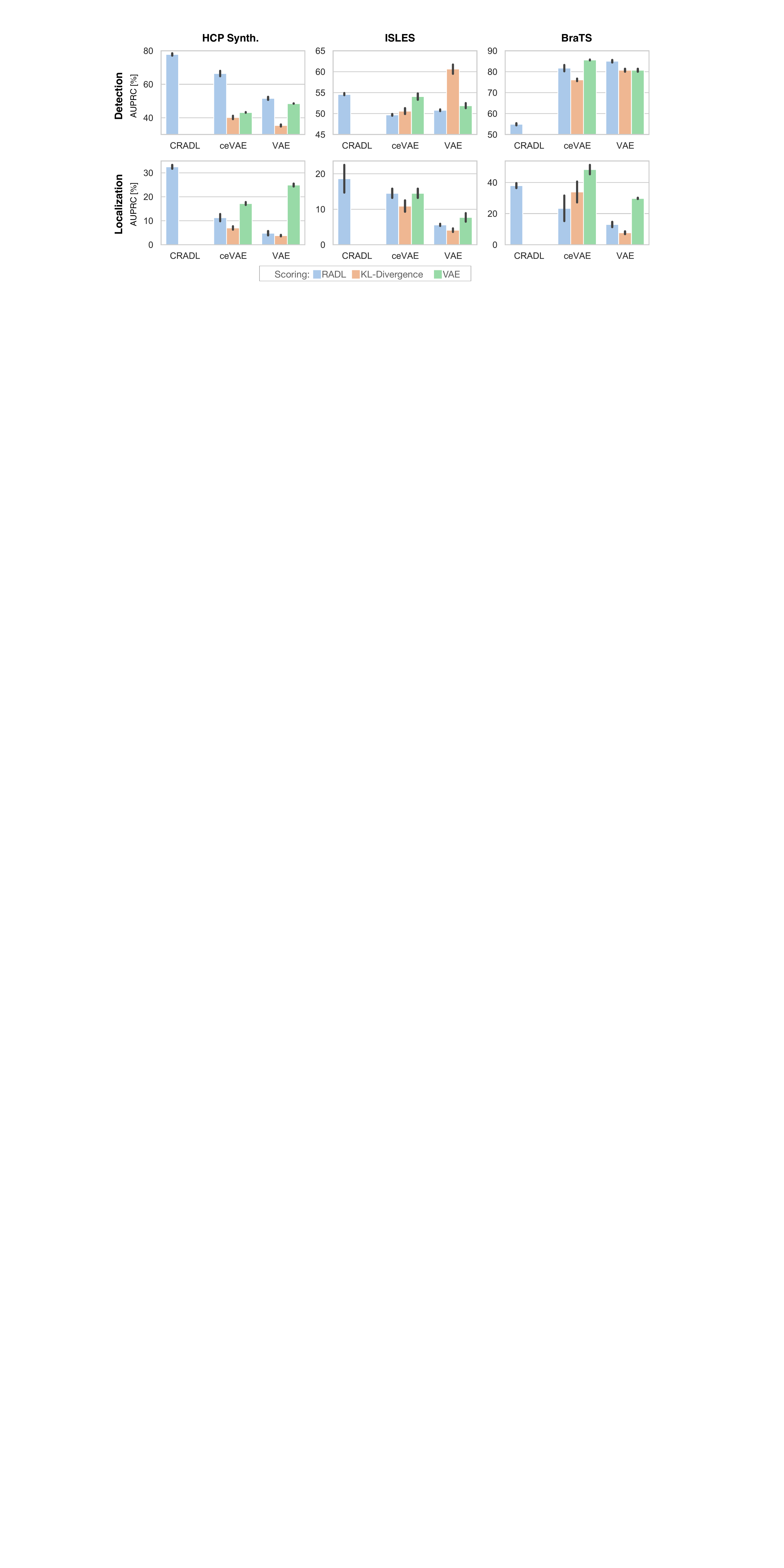}
    \caption{Performance comparison of CRADL with representation-based anomaly detection and localization (RADL) on the ceVAE and VAE alongside the KL-Divergence and the best VAE scoring method to assess the performance of purely representation-based approaches in general and the advantage of contrastive representations. 
    }
    \label{fig:representation_comparison}
\end{figure}
This experiment was performed to assess two things with regard to the down-stream tasks of anomaly detection and localization.
Firstly, whether purely representation-based anomaly scoring for the generative methods VAE and ceVAE offers advantages over standard anomaly scoring.
To this end, we define the representation-based anomaly detection and localization approach (\textbf{RADL}) where the use of a generative model on the representations of CRADL (see \cref{ss:methods-generative}) is emulated on the representations of a VAE and ceVAE.
Secondly, we want to assess whether the contrastive representations of CRADL are better suited for the two aforementioned down-stream tasks than the representations of generative models (generative pretext-task).

Tackling both questions, we depict both anomaly detection and localization performance in Fig. \ref{fig:representation_comparison}, where we followed the score selection criterion detailed in \cref{s:expsetup}.
Regarding the effectiveness of the purely representation-based RADL on the VAE and ceVAE, we compare it with the KL-Divergence and the best VAE scoring method selected in sections \ref{ss:results-detection} \& \ref{ss:results-localization}.
For Anomaly Localization it improves the performance over the KL-Divergence on BraTS and on HCP Synth it improves upon the best-performing VAE scoring method while underperforming on ISLES. 
For Anomaly Detection the RADL approach leads to substantial performance improvements over the KL-Divergence in all cases except for the ceVAE on BraTS, but it never outperforms the VAE scoring methods also using reconstruction. 

Based on these results we conclude that the RADL approach for generative methods using the representations yields performance benefits over the KL-Divergence and allows to detection of fine differences in the distribution as showcased by the results on HCP Synth..

Regarding the second point, assessing our hypothesis regarding performance benefits of the contrastive representations of CRADL, due to carrying more semantic information, compared with generative representations. 
We see our results as strengthening our hypothesis due to the clear performance improvements of CRADL for anomaly localization over the KL-Divergence and RADL approaches for anomaly localization and the performance improvement on HCP Synth. over the ceVAE and VAE performance as well as on ISLES with the exception of the KL-Divergence of the VAE. 
Even though, CRADL has the worst performance on BraTS for anomaly detection is directly counteracting our hypothesis. 
We argue again that this behavior might stem from the distribution shift already mentioned and that the contrastive representations allow for better detection and localization of fine semantic differences between anomalous and normal brain volumes as the HCP Synth. dataset provides
A further supporting fact of the advantage of a self-supervised pretext task is that the ceVAE, which also employs a self-supervised task, outperforms the VAE.

\subsection{Caveat of Experimental Setup}
Our experimental evaluations suffer from either being a synthetic problem for HCP synth. or the distribution shifts of BraTS and ISLES discussed in section \ref{ss:datasets}.
We believe that the experimental evidence on HCP synth. is the most trustworthy with regard to the specific task since structural changes due to age and unannotated anomalies in BraTS and ISLES do not allow showcasing the performance benefits of certain methods over others.
\footnote{Tumors and stroke lesions tend to glow brightly in T2-weighted MRI-Scans giving advantages to methods whose anomaly scores rule can be broken down to: "The more it glows, the more it is anomalous."}

Another critical aspect of the whole field, in general, is the high prevalence of anomalous cases during evaluation since this might lead to the possibility of methods being favored in evaluation metrics if giving per se high scores to certain regions of the brain prone to anomalies (here lesions). This aspect would need to be assessed either directly or evaluated by also having many normal images in the test datasets.

\section{Conclusion}
\label{s:conclusion}
In this work, we propose to purely use representations for unsupervised anomaly detection and localization and investigate to this end contrastive representations of a SimCLR-like pretext task. CRADL a simple framework for unsupervised anomaly detection and localization is the result of this investigation.

Concretely, we have shown that purely representation-based anomaly detection has benefits over generative models operating on image space and that representations of an encoder, trained with a contrastive pretext-task, in combination with a simple generative model, show advantageous properties over generative models on image space leading to better performance in the context anomaly detection and localization. 
This could clearly be observed for fine and diverse deviations from the distribution of normal samples.
For anomaly detection, slightly more powerful generative models like the normalizing flow operating on the representation space seem to perform best.
Whereas for anomaly localization, simpler generative models like a GMM perform better.

We believe the following two use cases make good use CRADL's properties for anomaly detection and localization:

\noindent - Cases with an Abundance of unannotated healthy and diseased images since the representation learning approach is decoupled from distribution learning. Further, it has been shown that using anomalous data for the pretext-task can even be helpful \cite{pinayaUnsupervisedBrainAnomaly2021} for a VQ-VAE.

\noindent -  Cases with a small healthy/normal dataset due to the low-dimensionality reducing the number of samples needed for the fit of the distribution with simpler generative models compared to generative modeling on image space.

To increase the potential use case for the medical domain a generalization to 3D inputs would be necessary and aggregation schemes for cases where a whole needs to be processed in smaller pieces s.a. the slice-wise approach here.
Further, since the approach of representation-based anomaly detection itself is very general as showcased by the experiments in \cref{ss:results-representations}, we believe that other self-supervised tasks \cite{li2021cutpaste,he2021masked} might further increase the performance of CRADL. 
Also, restorative techniques s.a. image restoration \cite{chenUnsupervisedLesionDetection2020a} and representation resampling \cite{pinayaUnsupervisedBrainAnomaly2021,marimont2021anomaly} might also be incorporated into the CRADL approach.

Anomaly detection and localization is a promising research field for the medical imaging domain, due to its potential to alleviate the annotation burden for medical screenings using deep learning to take place while having similar and understood properties to standard medical tests s.a. the blood count.
We hope that our proposed method and insights help push anomaly detection and localization into clinical practice.

\newpage

\section*{Acknowledgements}
Part of this work was funded by the Helmholtz Imaging Platform (HIP), a platform of the Helmholtz Incubator on Information and Data Science and the Helmholtz Association under the joint research school HIDSS4Health – Helmholtz Information and Data Science School for Health.
The authors would like to give a special thanks to Silvia Dias Almeida for not only helpful discussion and insights but also her unwavering support and believe in this work.


\printbibliography

@article{kellyKeyChallengesDelivering2019,
  title = {Key Challenges for Delivering Clinical Impact with Artificial Intelligence},
  author = {Kelly, Christopher J. and Karthikesalingam, Alan and Suleyman, Mustafa and Corrado, Greg and King, Dominic},
  year = {2019},
  month = dec,
  journal = {BMC Med},
  volume = {17},
  number = {1},
  pages = {195},
  issn = {1741-7015},
  doi = {10.1186/s12916-019-1426-2},
  langid = {english}
}

@article{mckinneyInternationalEvaluationAI2020,
  title = {International Evaluation of an {{AI}} System for Breast Cancer Screening},
  author = {McKinney, Scott Mayer and Sieniek, Marcin and Godbole, Varun and Godwin, Jonathan and Antropova, Natasha and Ashrafian, Hutan and Back, Trevor and Chesus, Mary and Corrado, Greg S. and Darzi, Ara and Etemadi, Mozziyar and {Garcia-Vicente}, Florencia and Gilbert, Fiona J. and {Halling-Brown}, Mark and Hassabis, Demis and Jansen, Sunny and Karthikesalingam, Alan and Kelly, Christopher J. and King, Dominic and Ledsam, Joseph R. and Melnick, David and Mostofi, Hormuz and Peng, Lily and Reicher, Joshua Jay and {Romera-Paredes}, Bernardino and Sidebottom, Richard and Suleyman, Mustafa and Tse, Daniel and Young, Kenneth C. and De Fauw, Jeffrey and Shetty, Shravya},
  year = {2020},
  month = jan,
  journal = {Nature},
  volume = {577},
  number = {7788},
  pages = {89--94},
  issn = {0028-0836, 1476-4687},
  doi = {10.1038/s41586-019-1799-6},
  langid = {english}
}

@book{bainDacieLewisPractical2017,
  title = {Dacie and {{Lewis}} Practical Haematology},
  editor = {Bain, Barbara J. and Bates, Imelda and Laffan, Michael A. and Lewis, S. M.},
  year = {2017},
  edition = {Twelfth edition},
  publisher = {{Elsevier}},
  address = {{Philadelphia, Pa.}},
  isbn = {978-0-7020-6925-3},
  langid = {english}
}

@article{aggarwal2021diagnostic,
  title={Diagnostic accuracy of deep learning in medical imaging: A systematic review and meta-analysis},
  author={Aggarwal, Ravi and Sounderajah, Viknesh and Martin, Guy and Ting, Daniel SW and Karthikesalingam, Alan and King, Dominic and Ashrafian, Hutan and Darzi, Ara},
  journal={NPJ digital medicine},
  volume={4},
  number={1},
  pages={1--23},
  year={2021},
  publisher={Nature Publishing Group}
}

@article{ZYTKOWSKI2021100105,
title = {Anatomical normality and variability: Historical perspective and methodological considerations},
journal = {Translational Research in Anatomy},
volume = {23},
pages = {100105},
year = {2021},
issn = {2214-854X},
doi = {https://doi.org/10.1016/j.tria.2020.100105},
url = {https://www.sciencedirect.com/science/article/pii/S2214854X20300443},
author = {Andrzej Żytkowski and R. Shane Tubbs and Joe Iwanaga and Edward Clarke and Michał Polguj and Grzegorz Wysiadecki},
}

@book{Goodfellow-et-al-2016,
    title={Deep Learning},
    author={Ian Goodfellow and Yoshua Bengio and Aaron Courville},
    publisher={MIT Press},
    note={\url{http://www.deeplearningbook.org}},
    year={2016}
}

@inproceedings{marimont2021anomaly,
  title={Anomaly detection through latent space restoration using vector quantized variational autoencoders},
  author={Marimont, Sergio Naval and Tarroni, Giacomo},
  booktitle={2021 IEEE 18th International Symposium on Biomedical Imaging (ISBI)},
  pages={1764--1767},
  year={2021},
  organization={IEEE}
}

@inproceedings{baur2018deep,
  title={Deep autoencoding models for unsupervised anomaly segmentation in brain MR images},
  author={Baur, Christoph and Wiestler, Benedikt and Albarqouni, Shadi and Navab, Nassir},
  booktitle={International MICCAI brainlesion workshop},
  pages={161--169},
  year={2018},
  organization={Springer}
}

@misc{he2021masked,
      title={Masked Autoencoders Are Scalable Vision Learners}, 
      author={Kaiming He and Xinlei Chen and Saining Xie and Yanghao Li and Piotr Dollár and Ross Girshick},
      year={2021},
      eprint={2111.06377},
      archivePrefix={arXiv},
      primaryClass={cs.CV}
}

@inproceedings{li2021cutpaste,
  author={Li, Chun-Liang and Sohn, Kihyuk and Yoon, Jinsung and Pfister, Tomas},
  booktitle={2021 IEEE/CVF Conference on Computer Vision and Pattern Recognition (CVPR)}, 
  title={CutPaste: Self-Supervised Learning for Anomaly Detection and Localization}, 
  year={2021},
  volume={},
  number={},
  pages={9659-9669},
  doi={10.1109/CVPR46437.2021.00954}}

@misc{meissen2021challenging,
      title={Challenging Current Semi-Supervised Anomaly Segmentation Methods for Brain MRI}, 
      author={Felix Meissen and Georgios Kaissis and Daniel Rueckert},
      year={2021},
      eprint={2109.06023},
      archivePrefix={arXiv},
      primaryClass={eess.IV}
}

@article{bakasAdvancingCancerGenome2017,
  title = {Advancing {{The Cancer Genome Atlas}} Glioma {{MRI}} Collections with Expert Segmentation Labels and Radiomic Features},
  author = {Bakas, Spyridon and Akbari, Hamed and Sotiras, Aristeidis and Bilello, Michel and Rozycki, Martin and Kirby, Justin S. and Freymann, John B. and Farahani, Keyvan and Davatzikos, Christos},
  year = {2017},
  month = dec,
  volume = {4},
  pages = {170117},
  issn = {2052-4463},
  doi = {10.1038/sdata.2017.117},
  journal = {Sci Data},
  language = {en},
  number = {1}
}

@article{baurAutoencodersUnsupervisedAnomaly2020,
author = {Baur, Christoph and Denner, Stefan and Wiestler, Benedikt and Navab, Nassir and Albarqouni, Shadi},
year = {2021},
month = {01},
pages = {101952},
title = {Autoencoders for unsupervised anomaly segmentation in brain MR images: A comparative study},
volume = {69},
journal = {Medical Image Analysis},
doi = {10.1016/j.media.2020.101952}
}

@article{chenBigSelfSupervisedModels2020,
  title = {Big {{Self}}-{{Supervised Models}} Are {{Strong Semi}}-{{Supervised Learners}}},
  author = {Chen, Ting and Kornblith, Simon and Swersky, Kevin and Norouzi, Mohammad and Hinton, Geoffrey},
  year = {2020},
  month = oct,
  archiveprefix = {arXiv},
  eprint = {2006.10029},
  eprinttype = {arxiv},
  journal = {arXiv:2006.10029 [cs, stat]},
  primaryclass = {cs, stat}
}

@article{chenSimpleFrameworkContrastive2020,
  title = {A {{Simple Framework}} for {{Contrastive Learning}} of {{Visual Representations}}},
  author = {Chen, Ting and Kornblith, Simon and Norouzi, Mohammad and Hinton, Geoffrey},
  year = {2020},
  month = jun,
  archiveprefix = {arXiv},
  eprint = {2002.05709},
  eprinttype = {arxiv},
  journal = {arXiv:2002.05709 [cs, stat]},
  primaryclass = {cs, stat}
}

@article{chenUnsupervisedLesionDetection2020a,
  title = {Unsupervised Lesion Detection via Image Restoration with a Normative Prior},
  author = {Chen, Xiaoran and You, Suhang and Tezcan, Kerem Can and Konukoglu, Ender},
  year = {2020},
  month = aug,
  volume = {64},
  pages = {101713},
  issn = {13618415},
  doi = {10.1016/j.media.2020.101713},
  journal = {Medical Image Analysis},
  language = {en}
}

@article{dempsterMaximumLikelihoodIncomplete1977,
  title = {Maximum {{Likelihood}} from {{Incomplete Data Via}} the {{{\emph{EM}}}} {{Algorithm}}},
  author = {Dempster, A. P. and Laird, N. M. and Rubin, D. B.},
  year = {1977},
  month = sep,
  volume = {39},
  pages = {1--22},
  issn = {00359246},
  doi = {10.1111/j.2517-6161.1977.tb01600.x},
  journal = {Journal of the Royal Statistical Society: Series B (Methodological)},
  language = {en},
  number = {1}
}

@article{dinhDensityEstimationUsing2017a,
  title = {Density Estimation Using {{Real NVP}}},
  author = {Dinh, Laurent and {Sohl-Dickstein}, Jascha and Bengio, Samy},
  year = {2017},
  month = feb,
  archiveprefix = {arXiv},
  eprint = {1605.08803},
  eprinttype = {arxiv},
  journal = {arXiv:1605.08803 [cs, stat]},
  primaryclass = {cs, stat}
}

@article{grillBootstrapYourOwn2020,
  title = {Bootstrap Your Own Latent: {{A}} New Approach to Self-Supervised {{Learning}}},
  shorttitle = {Bootstrap Your Own Latent},
  author = {Grill, Jean-Bastien and Strub, Florian and Altch{\'e}, Florent and Tallec, Corentin and Richemond, Pierre H. and Buchatskaya, Elena and Doersch, Carl and Pires, Bernardo Avila and Guo, Zhaohan Daniel and Azar, Mohammad Gheshlaghi and Piot, Bilal and Kavukcuoglu, Koray and Munos, R{\'e}mi and Valko, Michal},
  year = {2020},
  month = sep,
  archiveprefix = {arXiv},
  eprint = {2006.07733},
  eprinttype = {arxiv},
  journal = {arXiv:2006.07733 [cs, stat]},
  primaryclass = {cs, stat}
}

@article{leeSimpleUnifiedFramework2018,
  title = {A {{Simple Unified Framework}} for {{Detecting Out}}-of-{{Distribution Samples}} and {{Adversarial Attacks}}},
  author = {Lee, Kimin and Lee, Kibok and Lee, Honglak and Shin, Jinwoo},
  year = {2018},
  month = oct,
  archiveprefix = {arXiv},
  eprint = {1807.03888},
  eprinttype = {arxiv},
  journal = {arXiv:1807.03888 [cs, stat]},
  primaryclass = {cs, stat}
}

@article{liangEnhancingReliabilityOutofdistribution2020,
  title = {Enhancing {{The Reliability}} of {{Out}}-of-Distribution {{Image Detection}} in {{Neural Networks}}},
  author = {Liang, Shiyu and Li, Yixuan and Srikant, R.},
  year = {2020},
  month = aug,
  archiveprefix = {arXiv},
  eprint = {1706.02690},
  eprinttype = {arxiv},
  journal = {arXiv:1706.02690 [cs, stat]},
  primaryclass = {cs, stat}
}

@article{loshchilovSGDRStochasticGradient2017,
  title = {{{SGDR}}: {{Stochastic Gradient Descent}} with {{Warm Restarts}}},
  shorttitle = {{{SGDR}}},
  author = {Loshchilov, Ilya and Hutter, Frank},
  year = {2017},
  month = may,
  archiveprefix = {arXiv},
  eprint = {1608.03983},
  eprinttype = {arxiv},
  journal = {arXiv:1608.03983 [cs, math]},
  primaryclass = {cs, math}
}

@article{maaloeBIVAVeryDeep2019,
  title = {{{BIVA}}: {{A Very Deep Hierarchy}} of {{Latent Variables}} for {{Generative Modeling}}},
  shorttitle = {{{BIVA}}},
  author = {Maal{\o}e, Lars and Fraccaro, Marco and Li{\'e}vin, Valentin and Winther, Ole},
  year = {2019},
  month = nov,
  archiveprefix = {arXiv},
  eprint = {1902.02102},
  eprinttype = {arxiv},
  journal = {arXiv:1902.02102 [cs, stat]},
  primaryclass = {cs, stat}
}

@article{maierISLES2015Public2017,
  title = {{{ISLES}} 2015 - {{A}} Public Evaluation Benchmark for Ischemic Stroke Lesion Segmentation from Multispectral {{MRI}}},
  author = {Maier, Oskar and Menze, Bjoern H. and {von der Gablentz}, Janina and H{\"a}ni, Levin and Heinrich, Mattias P. and Liebrand, Matthias and Winzeck, Stefan and Basit, Abdul and Bentley, Paul and Chen, Liang and Christiaens, Daan and Dutil, Francis and et al.},
  year = {2017},
  month = jan,
  volume = {35},
  pages = {250--269},
  issn = {13618415},
  doi = {10.1016/j.media.2016.07.009},
  journal = {Medical Image Analysis},
  language = {en}
}

@article{nalisnickDeepGenerativeModels2019,
  title = {Do {{Deep Generative Models Know What They Don}}'t {{Know}}?},
  author = {Nalisnick, Eric and Matsukawa, Akihiro and Teh, Yee Whye and Gorur, Dilan and Lakshminarayanan, Balaji},
  year = {2019},
  month = feb,
  archiveprefix = {arXiv},
  eprint = {1810.09136},
  eprinttype = {arxiv},
  journal = {arXiv:1810.09136 [cs, stat]},
  primaryclass = {cs, stat}
}

@article{nalisnickDetectingOutofDistributionInputs2019,
  title = {Detecting {{Out}}-of-{{Distribution Inputs}} to {{Deep Generative Models Using Typicality}}},
  author = {Nalisnick, Eric and Matsukawa, Akihiro and Teh, Yee Whye and Lakshminarayanan, Balaji},
  year = {2019},
  month = oct,
  archiveprefix = {arXiv},
  eprint = {1906.02994},
  eprinttype = {arxiv},
  journal = {arXiv:1906.02994 [cs, stat]},
  primaryclass = {cs, stat}
}

@article{pinayaUnsupervisedBrainAnomaly2021,
  title = {Unsupervised {{Brain Anomaly Detection}} and {{Segmentation}} with {{Transformers}}},
  author = {Pinaya, Walter Hugo Lopez and Tudosiu, Petru-Daniel and Gray, Robert and Rees, Geraint and Nachev, Parashkev and Ourselin, Sebastien and Cardoso, M. Jorge},
  year = {2021},
  month = feb,
  archiveprefix = {arXiv},
  eprint = {2102.11650},
  eprinttype = {arxiv},
  journal = {arXiv:2102.11650 [cs, eess, q-bio]},
  primaryclass = {cs, eess, q-bio}
}

@article{radfordUnsupervisedRepresentationLearning2016,
  title = {Unsupervised {{Representation Learning}} with {{Deep Convolutional Generative Adversarial Networks}}},
  author = {Radford, Alec and Metz, Luke and Chintala, Soumith},
  year = {2016},
  month = jan,
  archiveprefix = {arXiv},
  eprint = {1511.06434},
  eprinttype = {arxiv},
  journal = {arXiv:1511.06434 [cs]},
  primaryclass = {cs}
}

@article{renLikelihoodRatiosOutofDistribution2019,
  title = {Likelihood {{Ratios}} for {{Out}}-of-{{Distribution Detection}}},
  author = {Ren, Jie and Liu, Peter J. and Fertig, Emily and Snoek, Jasper and Poplin, Ryan and DePristo, Mark A. and Dillon, Joshua V. and Lakshminarayanan, Balaji},
  year = {2019},
  month = dec,
  archiveprefix = {arXiv},
  eprint = {1906.02845},
  eprinttype = {arxiv},
  journal = {arXiv:1906.02845 [cs, stat]},
  primaryclass = {cs, stat}
}

@article{schleglFAnoGANFastUnsupervised2019,
  title = {F-{{AnoGAN}}: {{Fast}} Unsupervised Anomaly Detection with Generative Adversarial Networks},
  shorttitle = {F-{{AnoGAN}}},
  author = {Schlegl, Thomas and Seeb{\"o}ck, Philipp and Waldstein, Sebastian M. and Langs, Georg and {Schmidt-Erfurth}, Ursula},
  year = {2019},
  month = may,
  volume = {54},
  pages = {30--44},
  issn = {13618415},
  doi = {10.1016/j.media.2019.01.010},
  journal = {Medical Image Analysis},
  language = {en}
}

@article{vanessenHumanConnectomeProject2012,
  title = {The {{Human Connectome Project}}: {{A}} Data Acquisition Perspective},
  shorttitle = {The {{Human Connectome Project}}},
  author = {Van Essen, D.C. and Ugurbil, K. and Auerbach, E. and Barch, D. and Behrens, T.E.J. and Bucholz, R. and Chang, A. and Chen, L. and Corbetta, M. and Curtiss, S.W. and Della Penna, S. and Feinberg, D. and Glasser, M.F. and Harel, N. and Heath, A.C. and {Larson-Prior}, L. and Marcus, D. and Michalareas, G. and Moeller, S. and Oostenveld, R. and Petersen, S.E. and Prior, F. and Schlaggar, B.L. and Smith, S.M. and Snyder, A.Z. and Xu, J. and Yacoub, E.},
  year = {2012},
  month = oct,
  volume = {62},
  pages = {2222--2231},
  issn = {10538119},
  doi = {10.1016/j.neuroimage.2012.02.018},
  journal = {NeuroImage},
  language = {en},
  number = {4}
}

@article{winkensContrastiveTrainingImproved2020,
  title = {Contrastive {{Training}} for {{Improved Out}}-of-{{Distribution Detection}}},
  author = {Winkens, Jim and Bunel, Rudy and Roy, Abhijit Guha and Stanforth, Robert and Natarajan, Vivek and Ledsam, Joseph R. and MacWilliams, Patricia and Kohli, Pushmeet and Karthikesalingam, Alan and Kohl, Simon and Cemgil, Taylan and Eslami, S. M. Ali and Ronneberger, Olaf},
  year = {2020},
  month = jul,
  archiveprefix = {arXiv},
  eprint = {2007.05566},
  eprinttype = {arxiv},
  journal = {arXiv:2007.05566 [cs, stat]},
  primaryclass = {cs, stat}
}

@article{xiaoLikelihoodRegretOutofDistribution2020,
  title = {Likelihood {{Regret}}: {{An Out}}-of-{{Distribution Detection Score For Variational Auto}}-Encoder},
  shorttitle = {Likelihood {{Regret}}},
  author = {Xiao, Zhisheng and Yan, Qing and Amit, Yali},
  year = {2020},
  month = apr,
  archiveprefix = {arXiv},
  eprint = {2003.02977},
  eprinttype = {arxiv},
  journal = {arXiv:2003.02977 [cs, stat]},
  primaryclass = {cs, stat}
}

@article{zhangHybridModelsOpen2020,
  title = {Hybrid {{Models}} for {{Open Set Recognition}}},
  author = {Zhang, Hongjie and Li, Ang and Guo, Jie and Guo, Yanwen},
  year = {2020},
  month = aug,
  archiveprefix = {arXiv},
  eprint = {2003.12506},
  eprinttype = {arxiv},
  journal = {arXiv:2003.12506 [cs]},
  primaryclass = {cs}
}

@article{zimmererContextencodingVariationalAutoencoder2018a,
  title = {Context-Encoding {{Variational Autoencoder}} for {{Unsupervised Anomaly Detection}}},
  author = {Zimmerer, David and Kohl, Simon A. A. and Petersen, Jens and Isensee, Fabian and {Maier-Hein}, Klaus H.},
  year = {2018},
  month = dec,
  archiveprefix = {arXiv},
  eprint = {1812.05941},
  eprinttype = {arxiv},
  journal = {arXiv:1812.05941 [cs, stat]},
  primaryclass = {cs, stat}
}

@article{zimmererMedicalOutofDistributionAnalysis2020,
  title = {Medical {{Out}}-of-{{Distribution Analysis Challenge}}},
  author = {Zimmerer, David and Petersen, Jens and K{\"o}hler, Gregor and J{\"a}ger, Paul and Full, Peter and Ro{\ss}, Tobias and Adler, Tim and Reinke, Annika and {Maier-Hein}, Lena and {Maier-Hein}, Klaus},
  year = {2020},
  month = mar,
  publisher = {{Zenodo}},
  doi = {10.5281/ZENODO.3784230},
  copyright = {Creative Commons Attribution No Derivatives 4.0 International, Open Access}
}

@article{zimmererUnsupervisedAnomalyLocalization2019,
  title = {Unsupervised {{Anomaly Localization}} Using {{Variational Auto}}-{{Encoders}}},
  author = {Zimmerer, David and Isensee, Fabian and Petersen, Jens and Kohl, Simon and {Maier-Hein}, Klaus},
  year = {2019},
  month = jul,
  archiveprefix = {arXiv},
  eprint = {1907.02796},
  eprinttype = {arxiv},
  journal = {arXiv:1907.02796 [cs, eess, stat]},
  primaryclass = {cs, eess, stat}
}

\newpage
\appendix
\section{Results}
All results using all scoring methods for all anomaly detection and localization approaches are shown here.
In \cref{tab:model-select} the selected scoring methods shown in \cref{s:results} are displayed.
For slice-wise anomaly detection the validation and test set performance is shown in \cref{tab:samples_app} and \cref{tab:samples_test_app}.
For voxel-wise anomaly localization the validation and test set performance is shown in \cref{tab:pix_val_app} and \cref{tab:pix_test_app}.

\begin{table*}[h!]
    \centering
    \begin{tabular}{ll|cc|cc}
\toprule
\multirow{2}{*}{Dataset}    & \multirow{2}{*}{Method} & \multicolumn{2}{c|}{Anomaly Detection} & \multicolumn{2}{c}{Anomaly Localization} \\
                            &                         & Section 5.1       & Section 5.3       & Section 5.2         & Section 5.3        \\
\midrule
\multirow{3}{*}{HCP Synth.} & CRADL                   & Norm. Flow  & Norm. Flow  & GMM 8               & GMM 8              \\
                            & VAE                     & ELBO              & GMM 8             & combi               & GMM 8              \\
                            & ceVAE                   & ELBO              & Norm. Flow  & combi               & GMM 8              \\
\hline
\multirow{3}{*}{BraTS}      & CRADL                   & GMM 1             & GMM 1             & GMM 2               & GMM 2              \\
                            & VAE                     & KL-Div            & GMM 8             & rec                 & GMM 4              \\
                            & ceVAE                   & ELBO              & GMM 1             & combi               & GMM 1              \\
\hline
\multirow{3}{*}{ISLES}      & CRADL                   & GMM 1             & GMM 1             & GMM 1               & GMM 1              \\
                            & VAE                     & ELBO              & GMM 8             & combi               & GMM 1              \\
                            & ceVAE                   & ELBO              & GMM 1             & combi               & GMM 1             \\
\bottomrule
\end{tabular}
    \caption{Model Selection for all results shown in \cref{s:results} based on validation set performance shown in \cref{tab:samples_app,tab:pix_val_app}}
    \label{tab:model-select}
\end{table*}

\begin{table*}[h!]
    \centering
    \resizebox{\textwidth}{!}{
\begin{tabular}{|lll|cc|cc|cc|}
\toprule
                    &     &         &             \multicolumn{2}{c|}{HCP Synth.} & \multicolumn{2}{c|}{BraTS} & \multicolumn{2}{c|}{ISLES} \\
        Pre-Text & Gen. Model & Score &   AUPRC &    AUROC & AUPRC & AUROC & AUPRC & AUROC  \\
\midrule
        \multirow{8}{*}{VAE} & GMM  1 Comp 
                & NLL-grad &  0.0236$\pm$0.0011 &  0.8501$\pm$0.0035 &  0.1282$\pm$0.0123 &  0.8889$\pm$0.0023 &  \bfseries 0.0563$\pm$0.0033 &   0.86$\pm$0.0016 \\
            & GMM  2 Comp 
                & NLL-grad &  0.0206$\pm$0.0006 &  0.8436$\pm$0.0022 &  0.1095$\pm$0.0096 &  0.8808$\pm$0.0021 &  0.0471$\pm$0.0024 &  0.8523$\pm$0.0036 \\
            & GMM  4 Comp 
                & NLL-grad &   0.0348$\pm$0.004 &  0.8851$\pm$0.0054 &  \bfseries 0.1295$\pm$0.0173 &   0.8918$\pm$0.0038 &  0.0562$\pm$0.0057 &  0.8578$\pm$0.0035 \\
            & GMM  8 Comp 
                & NLL-grad &   \bfseries 0.048$\pm$0.0086 &   0.9019$\pm$0.007 &   0.1345$\pm$0.013 &  0.8949$\pm$0.0026 &  0.0584$\pm$0.0058 &  0.8605$\pm$0.0028 \\
            & Norm. Flow 
                & NLL-grad &    0.04$\pm$0.0066 &  0.8885$\pm$0.0077 & 0.0978$\pm$0.012 &   0.8666$\pm$0.009 &  0.0498$\pm$0.0098 &  0.8516$\pm$0.0048 \\
            \cline{2-9}
            & VAE 
                & combi &   \bfseries 0.2491$\pm$0.0063 &  \bfseries 0.9546$\pm$0.0005 &  0.2269$\pm$0.0328 &  0.9198$\pm$0.0038 &   \bfseries 0.077$\pm$0.0122 &  \bfseries 0.8745$\pm$0.0059 \\
            &     & KL-Div.-grad &  0.0373$\pm$0.0032 &  0.8657$\pm$0.0014 &  0.0772$\pm$0.0091 &   0.8446$\pm$0.011 &  0.0409$\pm$0.0047 &  0.8466$\pm$0.0081 \\
            &     & Rec. &   0.2101$\pm$0.003 &  0.9511$\pm$0.0003 & \bfseries 0.2976$\pm$0.0035 &  \bfseries 0.9248 $\pm$0.0006 &  0.0513$\pm$0.0001 &  0.8532$\pm$0.0023 \\
            \hline
\multirow{8}{*}{ceVAE} & GMM  1 Comp 
                & NLL-grad &  0.1072$\pm$0.0109 &     0.901$\pm$0.01 &  \bfseries0.2343$\pm$0.0816 &   0.9129$\pm$0.0159 &  \bfseries 0.0618 $\pm$0.0176 &     0.86$\pm$0.0089 \\
            & GMM  2 Comp 
                 & NLL-grad &  0.0757$\pm$0.0057 &  0.8952$\pm$0.0041 &   0.177$\pm$0.0426 &  0.9062$\pm$0.0097 &  0.0449$\pm$0.0093 &  0.8445$\pm$0.0068 \\
            & GMM  4 Comp 
                & NLL-grad &  0.0967$\pm$0.0156 &  0.9116$\pm$0.0085 &  0.1906$\pm$0.0184 &   0.9105$\pm$0.004 &  0.0511$\pm$0.0123 &   0.8456$\pm$0.006 \\
            & GMM  8 Comp 
                & NLL-grad &   \bfseries 0.1122$\pm$0.016 &  0.9223$\pm$0.0058 &   0.2068$\pm$0.033 &   0.9119$\pm$0.006 &  0.0612$\pm$0.0084 &  0.8488$\pm$0.0055 \\
            & Norm. Flow 
                & NLL-grad&  0.0606$\pm$0.0148 &  0.9008$\pm$0.0101 & 0.1072$\pm$0.0159 &  0.8756$\pm$0.0079 &  0.0304$\pm$0.0007 &  0.8157$\pm$0.0013\\
            \cline{2-9}
            & VAE 
                & combi &  \bfseries 0.1716$\pm$0.0146 &  \bfseries 0.9212$\pm$0.004 &   \bfseries 0.483$\pm$0.0299 &   \bfseries 0.9482$\pm$0.0032 &  \bfseries 0.1451$\pm$0.0125 &  \bfseries 0.8794$\pm$0.0022 \\
             & & KL-Div.-grad &  0.0702$\pm$0.0069 &  0.8586$\pm$0.0047 &   0.3394$\pm$0.067 &  0.9252$\pm$0.0163 &  0.1085$\pm$0.0163 &  0.8785$\pm$0.0059 \\
             & & Rec. &  0.0913$\pm$0.0023 &  0.9266$\pm$0.0017 &  0.4073$\pm$0.0389 &  0.9269$\pm$0.0074 &  0.0653$\pm$0.0044 &   0.8544$\pm$0.005 \\
            \hline
\multirow{8}{*}{CRADL} & GMM  1 Comp 
                & NLL-grad &  0.2263$\pm$0.0112 &  0.9664$\pm$0.0017 &  0.3341$\pm$0.0402 &  0.9357$\pm$0.0035 &  \bfseries 0.1859$\pm$0.0385 & \bfseries 0.8977$\pm$0.0033 \\
            & GMM  2 Comp 
                & NLL-grad &  0.2243$\pm$0.0125 &  0.9685$\pm$0.0017 &  \bfseries 0.3802$\pm$0.0163 &  \bfseries 0.9418$\pm$0.0009 &    0.1653$\pm$0.02 &  0.8955$\pm$0.0029 \\
            & GMM  4 Comp 
                & NLL-grad &  0.2875$\pm$0.0101 &  0.9741$\pm$0.0006 &  0.3383$\pm$0.0161 &  0.9384$\pm$0.0012 &  0.1441$\pm$0.0024 &   0.8935$\pm$0.003 \\
            & GMM  8 Comp 
                & NLL-grad &  \bfseries 0.3246$\pm$0.0076 & \bfseries 0.9779$\pm$0.0003  &  0.2908$\pm$0.0199 &  0.9309$\pm$0.0022 &  0.1257$\pm$0.0151 &  0.8906$\pm$0.0019 \\
            & Norm. Flow 
                & NLL-grad &    0.0924$\pm$0.0097 &  0.9397$\pm$0.0031  &  0.1362$\pm$0.0102 &  0.8736$\pm$0.0068 & 0.0393$\pm$0.0044 &  0.8213$\pm$0.0153\\
\bottomrule
\end{tabular}
    }
    \caption{Voxel-Wise anomaly localization metrics on test datasets. \textbf{Values} are shown in the Results \& Discussion section and selected based on the results of the best AUPRC scores on the validation set (\cref{tab:pix_val_app})}
    \label{tab:pix_test_app}
\end{table*}
\begin{table*}[h!] 
    \centering
    \resizebox{\textwidth}{!}{
\begin{tabular}{|lll |cc|cc|cc|}
\toprule
                &     &       &           \multicolumn{2}{c|}{HCP Synth.} & \multicolumn{2}{c|}{BraTS} & \multicolumn{2}{c|}{ISLES} \\
    Pre-Text & Gen. Model & Score &   AUPRC &    AUROC & AUPRC & AUROC & AUPRC & AUROC  \\
    \midrule
\multirow{8}{*}{VAE} & GMM 1   & NLL-grad & 0.0353$\pm$0.0056 &  0.8547$\pm$0.0025 &  0.0935$\pm$0.0047 &  0.8841$\pm$0.0022 &  \bfseries 0.0676$\pm$0.0149 &  0.8745$\pm$0.0095 \\
       & GMM 2   & NLL-grad & 0.0276$\pm$0.0018 &  0.8487$\pm$0.0031 &  0.0799$\pm$0.0046 &  0.8751$\pm$0.0038 &  0.0548$\pm$0.0102 &  0.8632$\pm$0.0067 \\
       & GMM 4   & NLL-grad & 0.0503$\pm$0.0071 &  0.8859$\pm$0.0043 &  \bfseries 0.1009$\pm$0.0131 &  0.8897$\pm$0.0058 &  0.0627$\pm$0.0064 &  0.8701$\pm$0.0032 \\
       & GMM 8   & NLL-grad & \bfseries 0.0774$\pm$0.0035 &  0.9088$\pm$0.0023 &  0.0925$\pm$0.0073 &  0.8875$\pm$0.0041 &  0.0624$\pm$0.0029 &  0.8805$\pm$0.0006 \\
       & Norm. Flow   &  NLL-grad & 0.0395$\pm$0.0008 &  0.8665$\pm$0.0021 &  0.0691$\pm$0.0036 &  0.8584$\pm$0.0052 &  0.0509$\pm$0.0091 &   0.863$\pm$0.0091 \\
      \cline{2-9}
      & \multirow{3}{*}{VAE} & combi  &  \bfseries 0.2945$\pm$0.0059 &  0.9527$\pm$0.0005 &  0.1842$\pm$0.0403 &  0.9171$\pm$0.0061 &   \bfseries 0.0894$\pm$0.012 &  0.8812$\pm$0.0051 \\
      &  & KL-Div.-grad  &  0.0735$\pm$0.0012 &  0.8771$\pm$0.0007 &  0.0677$\pm$0.0096 &  0.8553$\pm$0.0131 &   0.041$\pm$0.0009 &  0.8456$\pm$0.0044 \\
      & & Rec.  &  0.2081$\pm$0.0042 &  0.9426$\pm$0.0004 & \bfseries 0.2248$\pm$0.0077 &  0.9119$\pm$0.0015 &  0.0543$\pm$0.0082 &  0.8618$\pm$0.0056 \\
\hline
\multirow{8}{*}{ceVAE} & GMM 1   & NLL-grad &  0.1212$\pm$0.024 &  0.9057$\pm$0.0111 &  \bfseries 0.2313$\pm$0.0829 &  0.9181$\pm$0.0138 &  \bfseries 0.0609$\pm$0.0166 &  0.8633$\pm$0.0059 \\
       & GMM 2   & NLL-grad & 0.0741$\pm$0.0142 &  0.8989$\pm$0.0065 &  0.1611$\pm$0.0291 &  0.9126$\pm$0.0051 &  0.0385$\pm$0.0056 &   0.8436$\pm$0.005 \\
       & GMM 4   & NLL-grad & 0.1067$\pm$0.0132 &  0.9167$\pm$0.0039 &  0.1955$\pm$0.0214 &  0.9163$\pm$0.0024 &  0.0423$\pm$0.0058 &  0.8513$\pm$0.0033 \\
       & GMM 8   & NLL-grad & 0.1464$\pm$0.0285 &   0.9286$\pm$0.007 &  0.1841$\pm$0.0162 &   0.911$\pm$0.0002 &  0.0404$\pm$0.0054 &  0.8471$\pm$0.0044 \\
       & Norm. Flow   & NLL-grad & 0.0907$\pm$0.0144 &  0.9002$\pm$0.0081 &  0.0784$\pm$0.0144 &  0.8681$\pm$0.0088 &  0.0311$\pm$0.0017 &  0.8223$\pm$0.0094 \\
\cline{2-9}
       & \multirow{3}{*}{VAE} & combi  &  \bfseries 0.2183$\pm$0.0206 &  0.9148$\pm$0.0057 &   \bfseries 0.4321$\pm$0.005 &  0.9393$\pm$0.0038 &  \bfseries 0.1628$\pm$0.0242 &  0.8847$\pm$0.0042 \\
       & & KL-Div.-grad  &   0.096$\pm$0.0096 &  0.8655$\pm$0.0037 &    0.3337$\pm$0.04 &  0.9317$\pm$0.0078 &  0.0956$\pm$0.0278 &  0.8751$\pm$0.0092 \\
       & & Rec.  &  0.1163$\pm$0.0019 &  0.9117$\pm$0.0021 &  0.2884$\pm$0.0403 &  0.9068$\pm$0.0088 &   0.1321$\pm$0.029 &  0.8649$\pm$0.0041 \\
\hline
\multirow{5}{*}{VAE} & GMM 1   & NLL-grad & 0.3309$\pm$0.0064 &   0.969$\pm$0.0013 &  0.3102$\pm$0.0129 &  0.9364$\pm$0.0007 &  \bfseries 0.3279$\pm$0.0096 &  0.9342$\pm$0.0027 \\
       & GMM 2   & NLL-grad & 0.3184$\pm$0.0063 &  0.9688$\pm$0.0012 &  \bfseries 0.3748$\pm$0.0129 &   0.942$\pm$0.0016 &  0.3523$\pm$0.0034 &  0.9259$\pm$0.0034 \\
       & GMM 4   & NLL-grad & 0.3269$\pm$0.0058 &  0.9698$\pm$0.0012 &  0.3284$\pm$0.0095 &  0.9369$\pm$0.0014 &  0.3173$\pm$0.0246 &  0.9175$\pm$0.0032 \\
       & GMM 8   & NLL-grad & \bfseries 0.3354$\pm$0.0085 &   0.9725$\pm$0.001 &  0.3179$\pm$0.0091 &  0.9376$\pm$0.0025 &  0.2908$\pm$0.0242 &  0.9187$\pm$0.0017 \\
       & Norm. Flow & NLL-grad & 0.1176$\pm$0.004 &  0.9297$\pm$0.0043 &  0.1474$\pm$0.0152 &  0.8943$\pm$0.0052 &  0.1159$\pm$0.0235 &  0.8781$\pm$0.0123 \\
\bottomrule
\end{tabular}
    }
    \caption{Pixel-Wise anomaly localization metrics on validation datasets. \textbf{Values} show the scoring methods with the highest AUPRC scores.}
    \label{tab:pix_val_app}
\end{table*}

\begin{table*}[h!] 
    \centering
    \resizebox{\textwidth}{!}{
    \begin{tabular}{|lll|cc|cc|cc|}
\toprule
                &     &         &             \multicolumn{2}{c|}{HCP Synth.} & \multicolumn{2}{c|}{BraTS} & \multicolumn{2}{c|}{ISLES} \\
    Pre-Text & Gen. Model & Score &   AUPRC &    AUROC & AUPRC & AUROC & AUPRC & AUROC  \\
\midrule
    \multirow{8}{*}{VAE} & GMM 1 & NLL &   0.3710 $\pm$ 0.0083 &  0.7238 $\pm$ 0.0072 &  0.8254 $\pm$ 0.0079 &  0.8481 $\pm$ 0.0034 &   0.5121 $\pm$ 0.0050 &  0.7093 $\pm$ 0.0007 \\
             & GMM 2 & NLL &  0.3516 $\pm$ 0.0056 &   0.7229 $\pm$ 0.0060 &  0.8054 $\pm$ 0.0076 &  0.8351 $\pm$ 0.0041 &  0.4812 $\pm$ 0.0066 &   0.6783 $\pm$ 0.004 \\
             & GMM 4 & NLL &  0.4383 $\pm$ 0.0104 &  0.7748 $\pm$ 0.0054 &  0.8271 $\pm$ 0.0051 &  0.8467 $\pm$ 0.0029 &  0.4766 $\pm$ 0.0076 &  0.6824 $\pm$ 0.0081 \\
             & GMM 8 & NLL &  \bfseries 0.5164 $\pm$ 0.0087 &  0.7961 $\pm$ 0.0044 &  \bfseries 0.8495 $\pm$ 0.0062 &   0.8574 $\pm$ 0.0030 &  \bfseries 0.5077 $\pm$ 0.0024 &  0.7059 $\pm$ 0.0033 \\
             & Norm. Flow & NLL &  0.5422 $\pm$ 0.0129 &    0.8010 $\pm$ 0.0050 &  0.8293 $\pm$ 0.0089 &  0.8462 $\pm$ 0.0051 &  0.4882 $\pm$ 0.0024 &  0.6838 $\pm$ 0.0075 \\
             \cline{2-9}
             & VAE & ELBO  & \bfseries 0.4853 $\pm$ 0.0041 &  \bfseries 0.7941 $\pm$ 0.0010&   0.8546 $\pm$ 0.0010&  0.8647 $\pm$ 0.0007 &  \bfseries 0.5188 $\pm$ 0.0055 & \bfseries 0.7174 $\pm$ 0.0022 \\
             &           & KL-Div. &  0.3536 $\pm$ 0.0059 &   0.7294 $\pm$ 0.0020 &  \bfseries 0.8073 $\pm$ 0.0069 & \bfseries  0.8329 $\pm$ 0.0050 &  0.6068 $\pm$ 0.0124 &   0.7612 $\pm$ 0.0030 \\
             &           & Rec. &   0.4889 $\pm$ 0.004 &  0.7983 $\pm$ 0.0009 &  0.8521 $\pm$ 0.0008 &   0.8640 $\pm$ 0.0008 &  0.5126 $\pm$ 0.0065 &  0.7108 $\pm$ 0.0026 \\
       \hline
       \multirow{8}{*}{ceVAE} & GMM 1 & NLL &  0.4758 $\pm$ 0.0271 &  0.7893 $\pm$ 0.0113 &  \bfseries 0.8173 $\pm$ 0.0147 &  0.8456 $\pm$ 0.0079 &  \bfseries 0.4967 $\pm$ 0.0191 &  0.6857 $\pm$ 0.0083 \\
             & GMM 2 & NLL &  0.4892 $\pm$ 0.0371 &  0.7958 $\pm$ 0.0133 &  0.7518 $\pm$ 0.0286 &  0.8063 $\pm$ 0.0094 &   0.3924 $\pm$ 0.022 &  0.5836 $\pm$ 0.0176 \\
             & GMM 4 & NLL &  0.5287 $\pm$ 0.0371 &  0.8076 $\pm$ 0.0123 &  0.8013 $\pm$ 0.0268 &  0.8308 $\pm$ 0.0147 &  0.4098 $\pm$ 0.0186 &  0.5981 $\pm$ 0.0246 \\
             & GMM 8 & NLL &    0.5900 $\pm$ 0.0328 &  0.8233 $\pm$ 0.0101 &  0.7824 $\pm$ 0.0395 &  0.8211 $\pm$ 0.0175 &  0.3824 $\pm$ 0.0415 &  0.5696 $\pm$ 0.0463 \\
             & Norm. Flow & NLL &  \bfseries 0.6642 $\pm$ 0.0162 &  0.8452 $\pm$ 0.0091 &   0.7296 $\pm$ 0.043 &    0.7700 $\pm$ 0.0285 &  0.4005 $\pm$ 0.0185 &  0.5976 $\pm$ 0.0277 \\
             \cline{2-9}
             & VAE & ELBO  & \bfseries 0.4321 $\pm$ 0.0021 &  \bfseries 0.7691 $\pm$ 0.0010&  \bfseries 0.8556 $\pm$ 0.0019 & \bfseries 0.8653 $\pm$ 0.0009 &  \bfseries 0.5412 $\pm$ 0.0082 & \bfseries 0.7266 $\pm$ 0.0008 \\
             &           & KL-Div. &    0.4022 $\pm$ 0.0100 &  0.7547 $\pm$ 0.0046 &  0.7612 $\pm$ 0.0063 &  0.8024 $\pm$ 0.0033 &  0.5064 $\pm$ 0.0071 &  0.6967 $\pm$ 0.0088 \\
             &           & Rec. &  0.4294 $\pm$ 0.0036 &  0.7683 $\pm$ 0.0007 &  0.8563 $\pm$ 0.0023 &   0.8660 $\pm$ 0.0012 &   0.5467 $\pm$ 0.009 &  0.7279 $\pm$ 0.0003 \\
\hline
\multirow{5}{*}{CRADL} & GMM 1 & NLL &  0.6988 $\pm$ 0.0032 &  0.8679 $\pm$ 0.0024 &  \bfseries 0.8187 $\pm$ 0.0024 &  \bfseries 0.8264 $\pm$ 0.0010&  \bfseries 0.5488 $\pm$ 0.0049 & \bfseries 0.6927 $\pm$ 0.0003 \\
             & GMM 2 & NLL &  0.7043 $\pm$ 0.0044 &  0.8733 $\pm$ 0.0023 &  0.7913 $\pm$ 0.0026 &  0.7967 $\pm$ 0.0011 &  0.4891 $\pm$ 0.0048 &   0.6367 $\pm$ 0.0020 \\
             & GMM 4 & NLL &  0.7538 $\pm$ 0.0027 &  0.8856 $\pm$ 0.0019 &   0.7787 $\pm$ 0.0030 &  0.7867 $\pm$ 0.0012 &  0.4831 $\pm$ 0.0001 &   0.6369 $\pm$ 0.0030 \\
             & GMM 8 & NLL &  0.7598 $\pm$ 0.0034 &  0.8871 $\pm$ 0.0021 &  0.7949 $\pm$ 0.0034 &   0.8010 $\pm$ 0.0017 &  0.5028 $\pm$ 0.0038 &  0.6441 $\pm$ 0.0011 \\
             & Norm. Flow & NLL & \bfseries 0.7777 $\pm$ 0.0069 & \bfseries 0.8979 $\pm$ 0.0027 &  0.6883 $\pm$ 0.0145 &  0.7163 $\pm$ 0.0139 &  0.4427 $\pm$ 0.0185 &  0.6289 $\pm$ 0.0212 \\
\bottomrule
\end{tabular}
    }
    \caption{Slice-Wise anomaly detection metrics on test datasets. \textbf{Values} are shown in the Results \& Discussion section and selected based on the results of the best AUPRC scores on the validation set (\cref{tab:samples_app})}
    \label{tab:samples_test_app}
\end{table*}
\begin{table*}[h!] 
    \centering
    \resizebox{\textwidth}{!}{
    \begin{tabular}{|lll |cc|cc|cc|}
\toprule
                &             &    &    \multicolumn{2}{c|}{HCP Synth.} & \multicolumn{2}{c|}{BraTS} & \multicolumn{2}{c|}{ISLES} \\
    Pre-Text &  Gen. Model &  Score & AUPRC &    AUROC & AUPRC & AUROC & AUPRC & AUROC  \\
\midrule
    \multirow{8}{*}{VAE} & GMM 1 & NLL & 0.4214 $\pm$ 0.0057 &  0.7269 $\pm$ 0.0054 &  0.8068 $\pm$ 0.0162 &  0.8463 $\pm$ 0.0032 &  0.7408 $\pm$ 0.0065 &   0.8450 $\pm$ 0.0036 \\
    & GMM 2 & NLL & 0.4120 $\pm$ 0.0052 &  0.7269 $\pm$ 0.0053 &  0.7899 $\pm$ 0.0128 &  0.8273 $\pm$ 0.0019 &  0.7149 $\pm$ 0.0022 &   0.8130 $\pm$ 0.0015 \\
    & GMM 4 & NLL & 0.4479 $\pm$ 0.0065 &  0.7613 $\pm$ 0.0035 &  0.8117 $\pm$ 0.0147 &  0.8342 $\pm$ 0.0086 &  0.6913 $\pm$ 0.0167 &  0.8042 $\pm$ 0.0105 \\
    & GMM 8 & NLL & \bfseries 0.5236 $\pm$ 0.0105 &   0.7890 $\pm$ 0.0046 &  \bfseries 0.8304 $\pm$ 0.0156 &  0.8506 $\pm$ 0.0064 & \bfseries 0.7487 $\pm$ 0.0072 &  0.8353 $\pm$ 0.0041 \\
    & Norm. Flow & NLL & 0.5126 $\pm$ 0.0074 &    0.7800 $\pm$ 0.0091 &   0.8024 $\pm$ 0.0050 &  0.8321 $\pm$ 0.0031 &  0.7184 $\pm$ 0.0165 &  0.8149 $\pm$ 0.0043 \\
    \cline{2-9}
    & \multirow{3}{*}{VAE} & ELBO &  \bfseries 0.5724 $\pm$ 0.0035 &  0.8161 $\pm$ 0.0011 &  0.7871 $\pm$ 0.0026 &   0.8464 $\pm$ 0.0010 &  \bfseries 0.7013 $\pm$ 0.0166 &  0.8312 $\pm$ 0.0078 \\
    & & KL-Div. &  0.4304 $\pm$ 0.0025 &  0.7686 $\pm$ 0.0045 &  \bfseries 0.7967 $\pm$ 0.0122 &  0.8438 $\pm$ 0.0048 &  0.6818 $\pm$ 0.0162 &  0.8439 $\pm$ 0.0133 \\
    & & Rec. &  0.5717 $\pm$ 0.0031 &  0.8172 $\pm$ 0.0007 &  0.7803 $\pm$ 0.0037 &  0.8436 $\pm$ 0.0009 &  0.6905 $\pm$ 0.0204 &  0.8251 $\pm$ 0.0095 \\
\hline
    \multirow{8}{*}{ceVAE}& GMM 1 & NLL & 0.5333 $\pm$ 0.0198 &  0.7918 $\pm$ 0.0151 &  \bfseries 0.7935 $\pm$ 0.0051 &  0.8396 $\pm$ 0.0032 & \bfseries 0.6492 $\pm$ 0.0785 &  0.8115 $\pm$ 0.0249 \\
       & GMM 2  & NLL & 0.5365 $\pm$ 0.0322 &  0.7915 $\pm$ 0.0148 &  0.7345 $\pm$ 0.0302 &   0.7937 $\pm$ 0.0120 &  0.5069 $\pm$ 0.1066 &  0.7083 $\pm$ 0.0393 \\
       & GMM 4 &  NLL & 0.5731 $\pm$ 0.0370 &  0.8101 $\pm$ 0.0188 &   0.7999 $\pm$ 0.018 &  0.8246 $\pm$ 0.0103 &   0.5716 $\pm$ 0.0810 &   0.7510 $\pm$ 0.0268 \\
       & GMM 8 & NLL & 0.6099 $\pm$ 0.0299 &  0.8238 $\pm$ 0.0127 &  0.7657 $\pm$ 0.0396 &  0.8107 $\pm$ 0.0199 &  0.5103 $\pm$ 0.1109 &  0.7065 $\pm$ 0.0487 \\
       & Norm.Flow & NLL & \bfseries 0.6616 $\pm$ 0.0315 &  0.8372 $\pm$ 0.0153 &  0.6305 $\pm$ 0.0372 &  0.7233 $\pm$ 0.0324 &  0.5497 $\pm$ 0.1038 &   0.7210 $\pm$ 0.0509 \\
       \cline{2-9}
       & \multirow{3}{*}{VAE} & ELBO &   \bfseries 0.5360 $\pm$ 0.0019 &  0.8036 $\pm$ 0.0011 &  \bfseries 0.8016 $\pm$ 0.0035 &   0.8570 $\pm$ 0.0011 &  \bfseries 0.7297 $\pm$ 0.0191 &  0.8381 $\pm$ 0.0095 \\
       & & KL-Div. &  0.4563 $\pm$ 0.0107 &  0.7768 $\pm$ 0.0037 &  0.7722 $\pm$ 0.0069 &  0.8215 $\pm$ 0.0045 &  0.6165 $\pm$ 0.0198 &  0.8044 $\pm$ 0.0143 \\
       & & Rec. &   0.5340 $\pm$ 0.0021 &   0.8030 $\pm$ 0.0012 &  0.7973 $\pm$ 0.0068 &  0.8546 $\pm$ 0.0032 &   0.7270 $\pm$ 0.0192 &    0.8359 $\pm$ 0.0100 \\
\hline
    \multirow{5}{*}{CRADL} & GMM 1 & NLL &  0.7380 $\pm$ 0.0046 &  0.8775 $\pm$ 0.0023 & \bfseries 0.7357 $\pm$ 0.0064 &  0.7791 $\pm$ 0.0032 & \bfseries 0.7306 $\pm$ 0.0059 &  0.8356 $\pm$ 0.0021 \\
             & GMM 2 & NLL & 0.7442 $\pm$ 0.0042 &  0.8797 $\pm$ 0.0021 &  0.7017 $\pm$ 0.0051 &  0.7341 $\pm$ 0.0013 &  0.6417 $\pm$ 0.0018 &  0.7506 $\pm$ 0.0028 \\
             & GMM 4 & NLL & 0.7781 $\pm$ 0.0044 &  0.8901 $\pm$ 0.0017 &  0.6832 $\pm$ 0.0071 &   0.7245 $\pm$ 0.0050 &  0.6329 $\pm$ 0.0139 &  0.7416 $\pm$ 0.0084 \\
             & GMM 8 & NLL & 0.7803 $\pm$ 0.0045 &  0.8911 $\pm$ 0.0021 &   0.7056 $\pm$ 0.0070 &  0.7436 $\pm$ 0.0026 &   0.6359 $\pm$ 0.0050 &  0.7431 $\pm$ 0.0037 \\
             & Norm. Flow & NLL & \bfseries 0.8006 $\pm$ 0.0112 & 0.8975 $\pm$ 0.0043 &  0.5983 $\pm$ 0.0234 &  0.6845 $\pm$ 0.0209 &   0.6420 $\pm$ 0.0151 &  0.7716 $\pm$ 0.0059 \\
\bottomrule
\end{tabular}
    }
    \caption{Slice-Wise anomaly detection metrics on validation datasets. \textbf{Values} show the scoring methods with the highest AUPRC scores.}
    \label{tab:samples_app}
\end{table*}
\section{Baselines}
\paragraph{VAE} The Variational Autoencoder uses an Encoder (Enc) producing a distribution in the latent space $q(z|x)=\mathrm{Enc}(x)$and a Decoder (Dec) $\hat{x}=\mathrm{Dec}(z)$ mapping from the latent space to the image space.
In the latent space a prior $p(z)$ is  taken to which $q(z|x)$ should converge.
It is trained using a reconstructive loss (L1) between the original image and reconstruction $\hat{x}=\mathrm{Dec}(z)$ for $z\sim q(z|x)$ with the KL-Divergence weighed against each using the factor $\beta =1$. 
\begin{equation} 
\mathcal{L}_{\mathrm{VAE}}(x) = \mathcal{L}_{\mathrm{Rec}}(x, \hat{x}) + 
    \beta \cdot \mathcal{D}_{\mathrm{KL}}(q(z|x) ||p(z))
\end{equation}

\paragraph{ceVAE} The context encoding Variational Autoencoder is built identically to the VAE. Still, it has an additional term in the loss function corresponding to an image restoration where an image with a masked out area $x_{\mathrm{mask}}$ should be restored (using the VAE $\hat{x}_{\mathrm{restore}} = \mathrm{Dec}(\mathbb{E}_{z\sim q(z|x)}[z])$)to the original image $x$.
The corresponding loss is added with a factor $\gamma=1$ to the training objective of the VAE.
\begin{equation}
    \mathcal{L}_{\mathrm{ceVAE}}(x) = \mathcal{L}_{\mathrm{VAE}} + 
    \gamma \mathcal{L}_{\mathrm{Rec}}(x, \hat{x}_{\mathrm{restore}})
\end{equation}

\section{Scoring}
\begin{figure}[h!]
    \centering
    \includegraphics[width=0.5\textwidth]{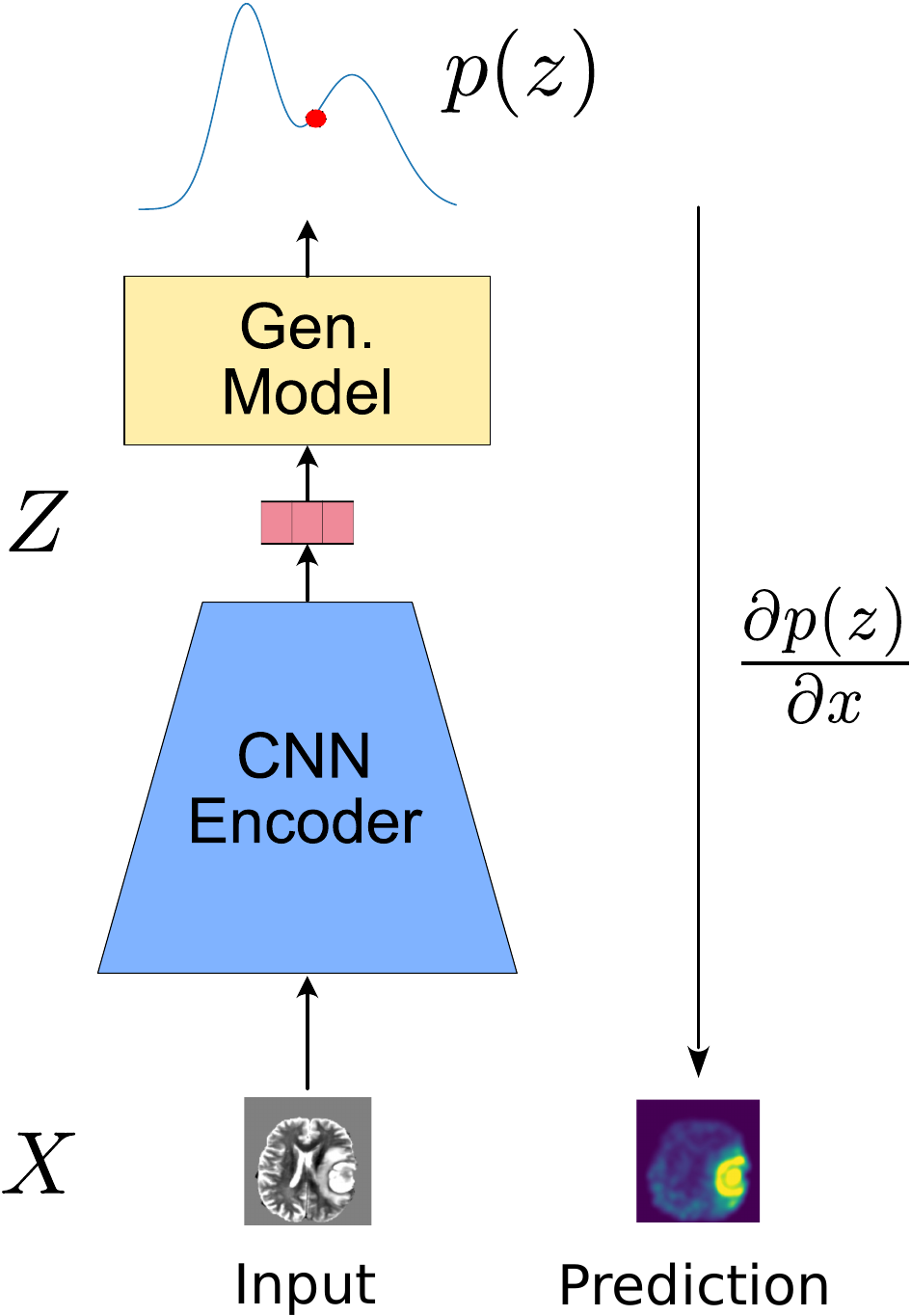}
    \caption{Visualization of the testing/ prediction phase of the model. The anomaly score in image space (anomaly heatmap) is calculated as the derivative of the predicted likelihood with respect to the input image.}
    \label{fig:pred_fig}
\end{figure}

All scoring methods for anomaly detection and anomaly localization are detailed in \cref{tab:pix_scores_app}.
A visualization of the inference phase of CRADL (and RADL) is detailed in \cref{fig:pred_fig}.

\begin{table*}[h!]
    \centering
    \begin{tabular}{l| l |l}
        Anomaly Task & Name & Formula \\
        \hline 
         Detection & NLL (proposed) & $s(x) = - \log(p(\mathrm{Enc}(x))$\\
        Detection & ElBO & $s(x) = \mathcal{L}_{\mathrm{VAE}}(x)$ \\
        Detection & KL-Div. & $s(x) = \mathcal{D}_{\mathrm{KL}}(q(z|x) ||p(z))$ \\
        Detection & Rec. & $s(x) = \mathcal{L}_{\mathrm{Rec}}(x, \hat{x}) $\\
        \hline
         Localization & nll-grad (proposed) &  $l_{\mathrm{nll-grad}}(x) = \left| \Delta_x (- \log (p(\mathrm{Enc}(x))) \right|$\\ 
         Localization & Rec. &  $l_{\mathrm{rec}}(x) = |x-\hat{x}|$ \\
         Localization & KL-Div.-grad \cite{zimmererContextencodingVariationalAutoencoder2018a} & $l_{\mathrm{kl-grad}}(x) =  \left|\Delta_x \mathcal{D}_{\mathrm{KL}}(q(z|x) ||p(z))
    \right|$ \\
        Localization & Combi \cite{zimmererContextencodingVariationalAutoencoder2018a} & $l_{\mathrm{combi}}(x) =  l_{\mathrm{kl-grad}}(x) \cdot l_{\mathrm{rec}}(x) $\\
    \end{tabular}
    \caption{Anomaly scoring methods for all anomaly detection and localization methods. For the VAE and ceVAE $\mathrm{Enc}(x)=\mathbb{E}_{z\sim(q(z|x)}[z]$ while for CRADL $\mathrm{Enc}(x)= f(x)$}
    \label{tab:pix_scores_app}
\end{table*}

\section{Experiment Details}
General details regarding the experiments is given below and in \cref{tab:DCAE} the design of the Deep-Convolutional Encoder and Decoder is shown.
\begin{itemize}
    \item Implementation: Pytorch and Pytorch Lightining.
    \item Hardware for Training: Single Nvidia GPUs with 12Gb VRAM (Titan XP, 2080Ti).
    \item Training times: SimCLR $\sim 12$h, GMM $\sim 1$m, RealNVP $\sim 30$m, VAE $\sim 24$h.
    \item Data augmentations implemented using batchgenerators\footnote{\url{https://github.com/MIC-DKFZ/batchgenerators}}:
    \begin{itemize}
        \item SimCLR: random mirroring, random cropping, random scaling, random multiplicative brightness, additive gaussian noise
        \item VAE: random scaling, random mirroring, rotations, multiplicative brightness, additive gaussian noise
    \end{itemize}
\end{itemize}
\begin{table*}[h!]
    \centering
    \begin{tabular}{l l}
    \toprule
    DC-Encoder & DC-Decoder\\
    \midrule
    Input $x$ & Input $z$ \\
    \hline
    4 x 4 Conv$_{\mathrm{nf}}$ Stride 2, BN, ReLU     &  
    4 x 4 Trans-Conv$_{\mathrm{16xnf}}$ Stride 1, BN, ReLU\\
    4 x 4 Conv$_{\mathrm{2xnf}}$ Stride 2, BN, ReLU & 
    4 x 4 Trans-Conv$_{\mathrm{8xnf}}$ Stride 2 Padding 1, BN, ReLU\\
    4 x 4 Conv$_{\mathrm{4xnf}}$ Stride 2, BN, ReLU & 
    4 x 4 Trans-Conv$_{\mathrm{4xnf}}$ Stride 2 Padding 1, BN, ReLU\\
    4 x 4 Conv$_{\mathrm{8xnf}}$ Stride 2. BN, ReLU& 
    4 x 4 Trans-Conv$_{\mathrm{2xnv}}$ Stride 2 Padding 1, BN, ReLU \\
    4 x 4 Conv$_{\mathrm{16xnf}}$ Stride 2, BN, ReLU & 
    4 x 4 Trans-Conv$_{\mathrm{nf}}$ Stride 2 Padding 1, BN, ReLU \\
    4 x 4 Conv$_{\mathrm{nz}}$ Stride 2 & 
    4 x 4 Trans-Conv$_{\mathrm{nc}}$ Stride 2 Padding 1, Sigmoid \\
    \hline
    Output $z$ & Output $\hat{x}$\\
    \bottomrule
    \end{tabular}
    \caption{Deep Convolutional Architecture, nf and nz are the hyperparameters for the architecture and bottleneck width. All experiments shown were conducted with nz=512 and nf=64. In the case of VAE and ceVAE nz$=512\cdot 2$
(BN: Batch Normalization)}
    \label{tab:DCAE}
\end{table*}

\paragraph{Gaussian Mixture Models} The Gaussian Mixture Models were implemented in Pytorch to enable fast fits and backpropagation of the NLL.

\paragraph{Normalizing Flow}
\begin{itemize}
    \item Architecture: We used the 8 consecutive flows of the Real NVP Architecture without splitting. Since the Model operates on the representations $z$ (dim=512), we use Linear layers inside the coupling blocks parametrized by $c_{\mathrm{block}}$=512 and $c_{\mathrm{out}}=c_{\mathrm{in}}=256$.
    \begin{itemize}
        \item Flow:
        \begin{enumerate}
            \item Coupling Block (Linear($c_{\mathrm{in}}$, $c_{\mathrm{block}}$), ReLU, Linear($c_{\mathrm{block}}, c_{\mathrm{out}}))$ 
            \item Random Permutation of Dimensions
        \end{enumerate}
    \end{itemize}
    \item Training Scheme:
    \begin{itemize}
        \item We added additional Gaussian noise ($\sigma=$1E-3) to our samples as preprocessing 
        \item Optimization Parameters: 60 epochs on the representations of the HCP training set with Adam Optimizer and the following parameters: learning rate 1E-3, weight decay 1E-4, gradient clipping (grad\_clip\_val=5 and additionally the learning rate is divided by 10 every 20 epochs
        \item We chose the model for later evaluation based on the smallest loss on the HCP validation set.
    \end{itemize}
\end{itemize}

\newpage

\section{Data Description}
\begin{figure*}[h!]
    \centering
    \includegraphics[width=0.95 \linewidth]{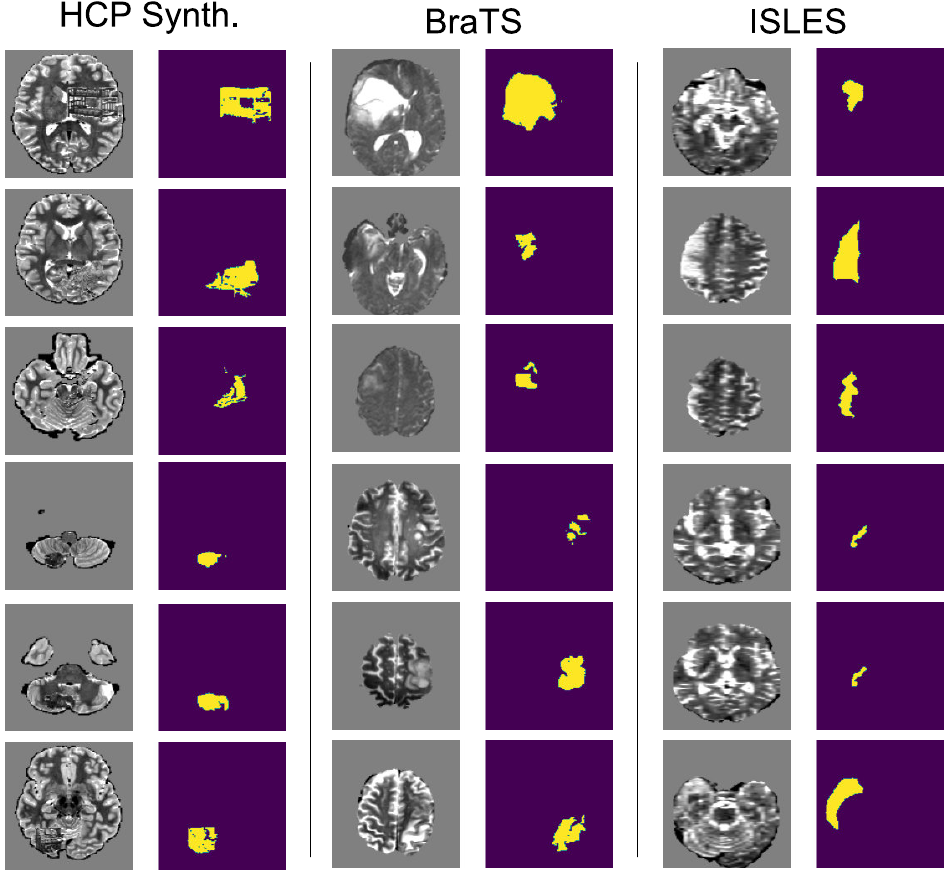}
    \caption{Visual examples of the anomalies present in our synthetically created in-house dataset HCP Synth., BraTS as well as ISLES. Shown are 6 exemplary images for each dataset with their corresponding reference annotations. Please note, that while for BraTS and ISLES, the anomaly primarily differs from the normal tissue by intensity (which can be seen as a bright area), for the HCP Synth. dataset the anomalies also differ by texture and hence might present a better use-case for anomaly detection methods \cite{meissen2021challenging}.}
    \label{fig:example_brains}
\end{figure*}

Additional to the statistical information about the dataset for evaulation in \cref{s:expsetup}, we also show visual examples in \cref{fig:example_brains}.

\end{document}